\documentclass[11pt,a4paper]{article}
\usepackage{times,latexsym}
\usepackage{url}
\usepackage[T1]{fontenc}
\usepackage{graphicx}
\usepackage{subcaption}

\usepackage{amsmath}
\usepackage{amssymb}
\usepackage{booktabs}
\usepackage{array}
\usepackage{colortbl} 
\usepackage{xcolor} 
\usepackage{adjustbox} 
\usepackage{hyperref}
\usepackage{multirow}
\usepackage{hyperref}
\usepackage{fontawesome5}

\usepackage[acceptedWithA]{tacl2021v1}

\usepackage{xspace,mfirstuc,tabulary}

\newif\iftaclinstructions
\taclinstructionsfalse 
\iftaclinstructions
\renewcommand{\confidential}{}
\renewcommand{\anonsubtext}{(No author info supplied here, for consistency with
TACL-submission anonymization requirements)}
\newcommand{\instr}
\fi

\iftaclpubformat 

\else

\fi

\title{\texttt{Lius:} Translation Model Based Instructional Lingustic \\ Using Continual Instruction Tuning In Kupang Malay}

\author{
  Joanito Agili Lopo$^1$, Yunita Sari$^2$, Guntur Budi Herwanto$^3$
  \\
  Department of Computer Science and Electronics
  \\
  Universitas Gadjah Mada
  \\
  $^1$\texttt{joanitoagililopo@mail.ugm.ac.id} \\
  $^2$\texttt{yunita.sari@ugm.ac.id}, 
  $^3$\texttt{gunturbudi@ugm.ac.id}
  \\
  \vspace{0.5em}
  \small{
  \href{https://huggingface.co/collections/joanitolopo/kupang-malay-dataset}
  {\faDatabase\ Dataset}
  \quad
  \href{https://huggingface.co/collections/joanitolopo/translation-models}
  {\faRobot\ Model}
  \quad
  \href{https://github.com/joanitolopo/instructional-linguistic-llm}
  {\faGithub\ Code}
  }
}

\date{}

\begin{document}
\maketitle
\begin{abstract}
Large Language Models (LLMs) offer new potential for translation tasks but often experience performance degradation when handling low-resource languages. To address this limitation, we propose an approach for fine-tuning LLMs on a low-resource language, Kupang Malay. Our approach involves designing a set of instructions by leveraging explicit lexical and semantic features from a bilingual dictionary, and introducing Continual Instruction Tuning (CIT), a training paradigm that enables iterative instruction-based training. Experimental results demonstrate that our model, named \texttt{Lius}, yields notable improvements over standard instruction-tuned models by outperforming 4–6 points, and surpassing both Neural Machine Translation (NMT) and Multilingual LLM models by 10–13 points on several evaluation metrics. These findings highlight the potential of our approach to mitigate the reliance on large-scale parallel data in low-resource language translation.
\end{abstract}

\section{Introduction}

Neural Machine Translation (NMT) represents a significant advancement in machine translation by leveraging Artificial Neural Networks (ANN) to model the conditional probability of a target sentence $y$ given a source sentence $x$. Architectures such as Multilayer LSTM \citep{NIPS2014_a14ac55a}, RNN Encoder-Decoder \citep{cho-etal-2014-learning}, Transformer \citep{NIPS2017_3f5ee243}, and Multilayer CNN \citep{gehring-etal-2017-convolutional} have demonstrated considerable improvements over earlier statistical approaches. However, NMT systems heavily depend on the availability of large-scale parallel corpora, which are predominantly available for high-resource languages like English, French, or Chinese. Meanwhile, many languages lack adequate parallel data and are often unsupported by commercial translation systems such as Google Translate \citep{hedderich-etal-2021-survey, haddow-etal-2022-survey}. 

The case of Indonesia illustrates this disparity. With over 718 identified local languages, only a small subset is represented in NLP resources. For instance, Javanese, despite having over 84 million speakers, is represented by merely 12 million parallel sentence pairs, contrast to Dutch, which has over 400 million in the OPUS \citep{tiedemann-2012-parallel} corpus. Initiatives such as NusaX and NusaWrites \citep{cahyawijaya-etal-2023-nusawrites, winata-etal-2023-nusax} have begun addressing these gaps, yet the collection of parallel data remains time-consuming and resource-intensive. Several challenges, such as a shortage of qualified annotators, dialectal variation, and inconsistent orthographic standards \citep{aji-etal-2022-one, novitasari-etal-2020-cross}, complicate both model development and parallel data collection efforts.

In response, researchers have explored monolingual corpora and bilingual dictionaries as alternatives. Monolingual data, which are more readily available, has been used to support unsupervised or semi-supervised NMT training, demonstrating notable improvements in output quality \citep{baziotis-etal-2020-language, chronopoulou-etal-2021-improving}. Bilingual dictionaries have also been utilized to replace rare or low-frequency words during training, leading to enhanced translation performance \citep{duan-etal-2020-bilingual, pourdamghani-etal-2019-translating}. Furthermore, multilingual training, transfer learning, and pivot-based strategies \citep{10.1145/3406095, leng-etal-2019-unsupervised, zoph-etal-2016-transfer} provide cross-lingual generalization capabilities that enhance adaptability and robustness, especially for low-resource settings. However, these approaches often lack the fine-grained lexical and semantic mappings available in parallel data and struggle to capture the complex linguistic phenomena in specific languages.

The rise of Large Language Models (LLMs) such as GPT-3 \citep{NEURIPS2020_1457c0d6} has significantly shifted the paradigm in machine translation. Trained on massive and diverse datasets in an unsupervised approach, LLMs exhibit strong generalization capabilities across various NLP tasks. These models are especially attractive for translation tasks as they are not strictly dependent on parallel corpora. Some studies have shown that, even in zero-shot settings, LLMs can perform competitively with traditional NMT systems trained on large parallel datasets \citep{lyu-etal-2024-paradigm}. Nonetheless, the performance of LLMs tends to degrade when applied to non-English or low-resource languages due to over-reliance on English during pretraining \citep{robinson-etal-2023-chatgpt, jiao2023chatgptgoodtranslatoryes, bang-etal-2023-multitask}. These models often fail to capture the rich lexical and semantic variation present in low-resource languages.

To address these limitations, this research adopts LLMs while proposing an Instructional Linguistics approach using Continual Instruction Tuning (CIT). The model will be trained continuously with linguistically informed instructions, incorporating lexical and semantic features from bilingual dictionaries, such as part-of-speech categories, synonyms, antonyms, and grammatical rules. This approach aims to enhance the model’s understanding of lexical-semantic relations, particularly in low-resource settings where such relations are rarely explicit. As a case study, this work focuses on Kupang Malay, a Malay-based creole spoken in the western part of Timor Island, which remains underrepresented in digital NLP resources.

\section{Related Work}
Recent studies have investigated various strategies to enhance multilingual translation models, particularly for low-resource languages, including approaches such as instruction tuning and prompt engineering. The following section outlines a recent approach implemented to address these challenges.

\paragraph{Instructiong Tuning}
Recent studies have demonstrated that leveraging monolingual and parallel corpora through approaches such as Low-Rank Adaptation (LoRA) on large language models, including Llama-2 \citep{touvron2023llama2openfoundation}, MaLA-500 \citep{lin2024mala500massivelanguageadaptation}, and Mistral \citep{jiang2023mistral7b}, enables significant improvement on the low-resource languages task, especially when pretraining is aligned with the linguistic characteristics of the target language \citep{iyer-etal-2024-exploring, lin2024mala500massivelanguageadaptation, jiang2023mistral7b}. Instruction tuning has also emerged as a key technique, where cross-lingual supervision is used to align bilingual or multilingual data in a way that enhances generalization across tasks and languages. This has been implemented through frameworks that extract structured instructions auch as word-level alignments using tools like FastAlign \citep{dyer_simple_2013}, and apply them to models such as BLOOMZ \citep{muennighoff-etal-2023-crosslingual} and XGLM \citep{lin2022fewshotlearningmultilinguallanguage}, achieving measurable improvements in translation quality, often surpassing previous baselines by up to several BLEU points \citep{cahyawijaya-etal-2023-instructalign, mao-yu-2024-tuning, li_eliciting_2024, lin2022fewshotlearningmultilinguallanguage, muennighoff-etal-2023-crosslingual, dyer_simple_2013}. 

\paragraph{Prompt Engineering}
\label{sec:prompt-engineering}
Instead of relying on a heavy training process, some studies have designed instruction-specific approaches for low-resource language translation, aiming to improve model interpretability and output quality. By combining monolingual corpora, bilingual dictionaries, and syntactic patterns, models like BLOOMZ \citep{muennighoff-etal-2023-crosslingual} and ChatGPT benefit from reduced ambiguity and improved lexical alignment \citep{guo-etal-2024-teaching, muennighoff-etal-2023-crosslingual}. Similar gains have been observed through the use of semantic embeddings such as LASER \citep{heffernan-etal-2022-bitext} to refine word correspondences in specific languages \citep{merx-etal-2024-low, heffernan-etal-2022-bitext}. Advances in prompt design, including domain-specific, morphologically-informed, and context-aware instructions, have shown to further support model reasoning and translation accuracy across tasks and models like GPT-4 and Mistral \citep{peng-etal-2023-towards, zhang-etal-2024-hire, huang-etal-2023-languages, he-etal-2024-exploring, jiang2023mistral7b}.

\section{Kupang Malay Language}
Kupang Malay is a Malay-based creole spoken in the western part of Timor Island and is widely used as a lingua franca across various ethnic communities. Although it originated through creolization rather than as a native ethnic language, it has evolved into a stable and distinct linguistic system \citep{Rafael2019}. The language exhibits minimal morphological complexity such as limited to four prefixes, with no suffixes or infixes, and a vowel system similar to Indonesian, excluding the schwa vowel \citep{jacob2003kamus}. Several dialects such as Air Mata, Alor Malay, and Basa Kupang reflect the diverse vernacular influences of its speakers.

Kupang Malay is classified as a low-resource language in the field of natural language processing (NLP). It lacks substantial parallel corpora in major multilingual datasets such as OPUS \citep{tiedemann-2012-parallel}, WikiMatrix, and CCMatrix \citep{schwenk-etal-2021-wikimatrix, schwenk-etal-2021-ccmatrix}. Existing resources for Kupang Malay are sparse and include only a few datasets such as Taxi1500 \citep{ma2024taxi1500multilingualdatasettext}, PanLex \citep{kamholz-etal-2014-panlex}, and the Bhinneka Korpus \citep{lopo2024constructingexpandinglowresourceunderrepresented}, each providing limited data coverage for the language. Despite being spoken by an estimated 5.3 million people in East Nusa Tenggara (NTT), Kupang Malay has yet to receive meaningful technological support. Bridging this gap through NLP development could not only promote digital inclusion for its speakers but also serve as a gateway to revitalizing over 70 other underrepresented languages in the region.

\section{Instructional Linguistic}
\label{sec:instructional_linguistic}
Instructional Linguistics is an approach that applies linguistic theories and principles to design an instruction to guide the model during training. There are four prompts that have been derived, i.e., Context-based \S\ref{sub:context-based}, Semantic-based \S\ref{sub:semantic-mapping-based}, Phonetic-based \S\ref{sub:phonetic-base}, and List-Group-Label-based \S\ref{sub:lgl-based}. Following section will explain in a details what is the theory behind this and the background of the created prompts.

\subsection{Background} 
In this study, Instructional Linguistics are motivated by Direct Instruction, introduced by Engelmann and Wesley Becker, where learning is delivered explicitly and systematically by the instructor \citep{Zahriani2014}. It is also aligned with Explicit Instruction, which emphasizes clear learning objectives and reduced cognitive load \citep{Hughes2017}. From a linguistic perspective, this approach resonates with Stephen D. Krashen’s Second Language Acquisition theory, which underscores the importance of comprehensible and sufficiently challenging input in the language learning process \citep{Pauzan2024}.

\subsection{Sentence Representation}

\begin{figure}[!t]
\includegraphics[width=\linewidth,trim={0, 0, 0, 0}, clip]{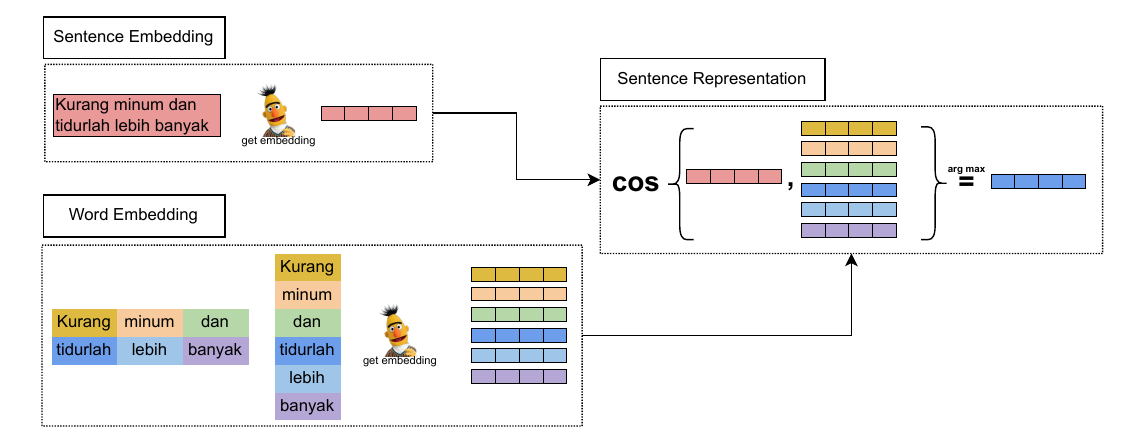}
\caption{Illustration of the sentence representation extraction.}
\label{fig:kata_kunci}
\end{figure}

The process of instruction generation requires a sentence representation that serves as a proxy for extracting lexical and semantic features relevant to the instruction. We employ the KeyBERT method \citep{grootendorst_maartengrkeybert_2021,201908.0073}\footnote{\url{https://github.com/MaartenGr/keyBERT}} using Indonesian BERT model \footnote{\url{https://huggingface.co/indobenchmark/indobert-large-p2}} \citep{wilie-etal-2020-indonlu} to identify the most semantically relevant word in a sentence. Figure \ref{fig:kata_kunci} illustrates the process of extraction of sentence representations for an Indonesian sentence. 

In general, the process begins by obtaining the sentence embedding representation $W_{id} = W_1,\ W_2,\ \ldots,W_m$, where $W_i$ represents each word in the sentence. The sentence embedding representation $E_{W_{id}}$ is derived from the special [CLS] token output by the BERT model, while the embedding of each word $E_{W_i}$ is obtained from its respective token representation. Next, cosine similarity is computed between the document embedding $E_{W_{id}}$ and each word embedding $E_{W_i}$ using Equation \ref{eq:bert_representation}.
\begin{equation}
\cos \left(E_{W_{id}},\ E_{W_i}\right) = \frac{E_{W_{id}} \cdot E_{W_i}}{\lVert E_{W_{id}} \rVert \lVert E_{W_i} \rVert}
\label{eq:bert_representation}
\end{equation}
To determine the most appropriate $W_i$ for each sentence, the word with the highest cosine similarity value is selected as the candidate for sentence representation, as defined in Equation \ref{eq:cos_rep}.

\begin{equation}
W^\ast = \arg\max\limits_{i} \cos \left(E_{W_{id}},\ E_{W_i}\right)
\label{eq:cos_rep}
\end{equation}

\subsection{Context-Based Prompt}
\label{sub:context-based}
The context-based prompt, inspired by the work of \citet{guo-etal-2024-teaching} and \citet{merx-etal-2024-low}, incorporates sentence examples in the target language that contain the semantic representation of the input sentence. Specifically, given an input sentence in Indonesian $S_{id}$ and its corresponding semantic representation $W^\ast$, the equivalent term in Kupang Malay, denoted as $W_{mkn}^\ast$, is retrieved using a bilingual dictionary.  However, if no equivalent $W_{mkn}^\ast$ is found, the original $W^\ast$ is retained without translation. Subsequently, the context-based instruction retrieves $n$ example sentences from the bilingual dictionary that contain the word $W_{mkn}^\ast$. These example sentences, ${S_{mkn}^1,\ S_{mkn}^2,\ldots,S_{mkn}^n}$, serve to reinforce the model’s understanding of the relationship between the source sentence representation and its contextual realization in the target language. 

\subsection{Semantic Mapping-Based Prompt} 
\label{sub:semantic-mapping-based}
A Semantic-based prompt inspired by the work of \citet{ghazvininejad2023dictionarybasedphraselevelpromptinglarge}, which provides bilingual word pairs in the prompt.
\begin{figure}[!t]
\includegraphics[width=\linewidth,trim={0, 0, 0, 0}, clip]{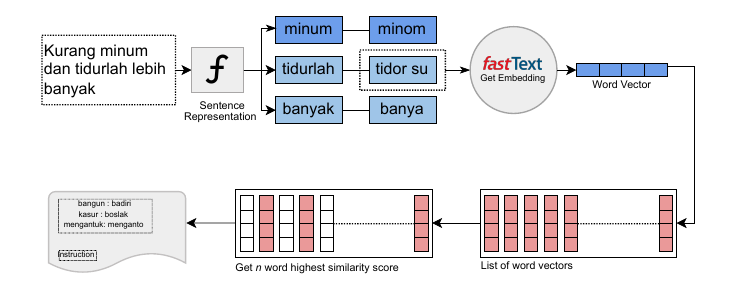}
\caption{Illustration of semantic mapping-based instruction.}
\label{fig:semantik_mapping}
\end{figure}
Firstly, the input $S_{id}$ is represented by a salient word $W^\ast$. This word is then searched for the parallel in Kupang Malay as $W_{mkn}^\ast$ using a bilingual dictionary. Next, $W_{mkn}^\ast$ is mapped to its vector representation $E(W_{mkn}^\ast)$ using FastText embeddings \citep{joulin-etal-2017-bag} trained on monolingual Kupang Malay text. The model then searches for the $n$ nearest neighboring words in Kupang Malay based on cosine similarity between $E(W_{mkn}^\ast)$ and all other word vectors $E(W_{mkn})$ in the target language. The retrieved nearest neighbors, ${W_{mkn_1}^\ast,\ W_{mkn_2}^\ast,\ldots,W_{mkn_i}^\ast}$, are then represented in Kupang Malay. Each of these words is subsequently matched with their Indonesian counterparts using the bilingual dictionary. An illustration of the construction is shown in Figure~\ref{fig:semantik_mapping}.

\subsection{Phonetic-Based Prompt}
\label{sub:phonetic-base}
Phonetic-based prompt enables the model to map sentence representations from the source language to the target language by identifying phonetically similar words, which is expected to improve model performance. This idea is motivated by the work of \citet{Atkinson1975} in the pedagogy field. 

Given Indonesian character $c_{id}^i$ at position $i$ and the Kupang Malay character $c_{mkn}^i$ at the same position, the instruction begins by finding the phonetic representation of the word $W_{mkn}^\ast$ and applying the phonetic rule $R$ (see Equation \ref{eq:transform_phonetic}) extracted from the bilingual dictionary. 
\begin{figure}[!t]
    \centering
    \includegraphics[width=\linewidth]{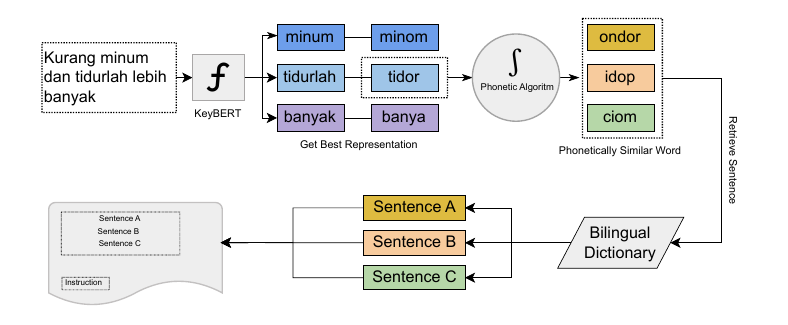}
    \caption{Illustration of phonetic-based instruction construction.}
    \label{fig:keyword_inst}
\end{figure}

\begin{equation}
R\left(c_{id}^i\right)\rightarrow\ c_{mkn}^i\ \forall i,\ if\ c_{id}^i\neq c_{mkn}^i
\label{eq:transform_phonetic}
\end{equation}
After the phonetic rule $R$ is obtained, the phonetic representation of the sentence $W_{mkn}^\ast$ is derived as \[
R(W_{mkn}^\ast) = \{R(c_1),\ R(c_2),\ldots,\ R(c_n)\}
\] where $\{c_1,\ c_2,\ldots,\ c_n\}$ are the characters in the word. To compute the phonetic similarity between $W_{mkn}^\ast$ and a candidate word $W_{mkn}$ in the target language, a sequence matching algorithm\footnote{\url{https://github.com/python/cpython/blob/3.13/Lib/difflib.py}} is used. The similarity score is calculated using Equation~\ref{eq:sequence_matching}, where $L$ is the length of the longest matching subsequence between $R(W_{mkn}^\ast)$ and $R(W_{mkn})$. If the similarity score $sim(W_{mkn}^\ast,\ W_{mkn}) \geq \tau$ (with threshold $\tau = 0.7$), the two words are considered phonetically similar. An illustration of construction is shown in Figure~\ref{fig:keyword_inst}.

\begin{equation}
d = \max\left(\text{len}(R(W_{mkn}^\ast)),\ \text{len}(R(W_{mkn}))\right)
\label{eq:phonetic_denom}
\end{equation}

\begin{equation}
sim(W_{mkn}^\ast,\ W_{mkn}) = \frac{L}{d}
\label{eq:sequence_matching}
\end{equation}


\subsection{List-Group-Label-Based Prompt} 
\label{sub:lgl-based}
This prompt is inspired by \citet{77f9438c-000e-36f6-978c-a8403f5ae833}, which introduces technical vocabulary through the categorization of sentences or objects within instructional content. The prompt generation begins by identifying a list of words related to a specific term, then grouping these words and assigning a label to each group. 

\begin{figure}[!t]
\includegraphics[width=\linewidth,trim={0, 0, 0, 0}, clip]{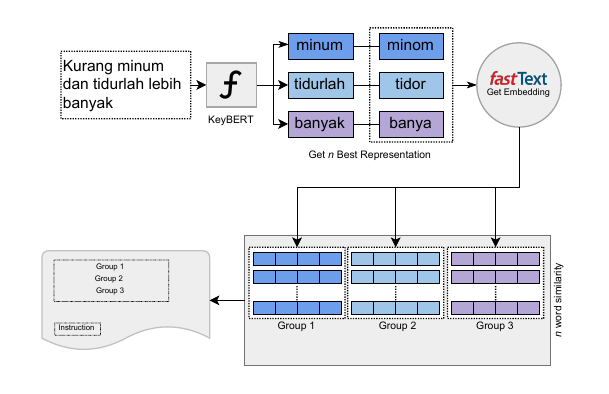}
\caption{Illustration of List-Group-Label based instruction.}
\label{fig:lgl_inst}
\end{figure}

Specifically, given an input sentence representation $W^\ast={W_1^\ast,\ W_2^\ast,\ \ldots,\ W_n^\ast}$, each representation $W_n^\ast$ is used to retrieve $n$ closest words ${W_{i1},\ W_{i2},\ \ldots,\ W_{in}}$ in the Kupang Malay language using FastText embeddings. 
This approach enriches the model with a broader understanding of vocabulary and context, especially when learning cross-lingual semantics. The illustration of this technique is illustrated in Figure~\ref{fig:lgl_inst}.

\section{Continual Instruction Tuning (CIT)}
\begin{figure}[!t]
    \includegraphics[width=\linewidth,trim={0, 0, 0, 0}, clip]{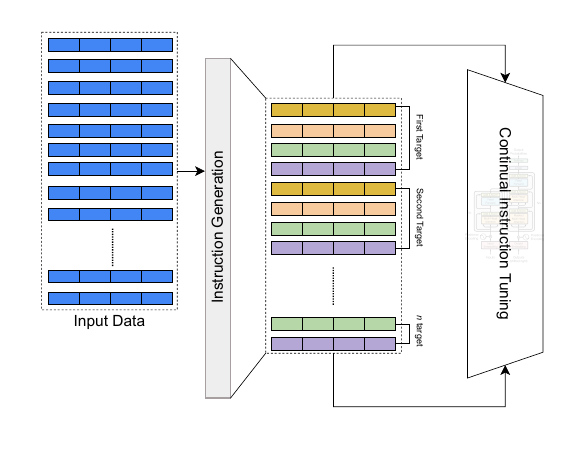}
    \caption{Continual Instruction Tuning adopts a continual training paradigm in which the model learns iteratively, processing one data sample for every four instruction prompts.}
    \label{fig:cit_framework}
\end{figure}

Continual Instruction Tuning (CIT) is a training paradigm proposed in this study to optimize the ability of Large Language Models (LLMs) in processing and leveraging our Instructional Linguistic \S\ref{sec:instructional_linguistic} approach. CIT draws inspiration from the work of \citet{lee-etal-2024-instruction} and \citet{feng2023citinglargelanguagemodels}, who explored LLM training through curriculum learning, rubrics, and self-correction mechanisms. 
Unlike conventional fine-tuning, which typically involves a single input–target pair, CIT is designed to continuously incorporate multiple instruction forms for the same input. Specifically, each input $x$ is associated with a set of instructions ${I_1, I_2, I_3, I_4}$, all corresponding to the same translation target $y$. The training process is carried out sequentially, where the model is fine-tuned with the pairs $(x, I_1) \rightarrow y$, followed by $(x, I_2) \rightarrow y$, and so on, until $(x, I_4) \rightarrow y$. Overview of the CIT-based fine-tuning process is illustrated in Figure~\ref{fig:cit_framework}.



\section{Experimental Settings}
\subsection{Data Acquisition}
\paragraph{Parallel Data}
The parallel sentences from Indonesian to Kupang Malay were collected and processed from several sources, namely the Bhinneka Korpus \citep{lopo2024constructingexpandinglowresourceunderrepresented}, BibleNLP\footnote{\url{https://github.com/BibleNLP/ebible}}, and The Language Archive\footnote{\url{https://archive.mpi.nl/tla/}}. A total of 66{,}521 sentence pairs were compiled, with 53{,}217 used for training and 13{,}304 for testing. Since the translation direction in this study is from Indonesian to Kupang Malay, English sentences in BibleNLP and The Language Archive were first translated into Indonesian using the NLLB model\footnote{\url{https://huggingface.co/facebook/nllb-200-distilled-1.3B}}. To ensure translation quality, back-translation was performed and evaluated using SacreBLEU and BERTScore\footnote{\url{https://huggingface.co/FacebookAI/roberta-large}}. The statistics of the parallel data and the back-translation results are presented in Table~\ref{tab:statistik_data}. In addition, a bilingual dictionary was compiled from Kamus Pengantar Bahasa Melayu Kupang by \citet{jacob2003kamus}, consisting of approximately 3{,}200 lexical entries. 

\begin{table}[!h]
\centering
\resizebox{\linewidth}{!}{
    \begin{tabular}{@{}l r r r@{}}
        \toprule
        \textbf{Source} & \textbf{\# Sentences} & \textbf{SacreBLEU} & \textbf{BERTScore} \\
        \midrule
        Bible NLP & 1,229 & 35.96 & 0.94 \\
        Bhinneka Corpus & 4,000 & -- & -- \\
        The Language Archive & 56,740 & 32.69 & 0.93 \\
        Tapaleuk & 4,552 & 24.34 & 0.90 \\
        \midrule
        \textbf{Total Parallel Data} & \textbf{66,521} & -- & -- \\
        \quad Train & 53,217 & -- & -- \\
        \quad Test & 13,304 & -- & -- \\
        \midrule
        \textbf{Total Instruction Data} & \textbf{227,172} & -- & -- \\
        \quad Train & 212,868 & -- & -- \\
        \quad Test & 13,304 & -- & -- \\
        \quad Experience Replay & 1,000 & -- & -- \\
        \bottomrule
    \end{tabular}
    }
\caption{Data statistics and back-translation evaluation. For \textbf{test instruction data} is randomly chosen over four instructional linguistic during testing.}
\label{tab:statistik_data}
\end{table}

\paragraph{Monolingual Data}
This study also employs monolingual data in Kupang Malay to support the development of a FastText model, which plays a crucial role in the instruction construction process within the instructional linguistic approach. The monolingual dataset is collected from several primary sources in Kupang Malay, including Tapaleuk News, the Jakarta Field Station \citep{jakarta_field_station}, Taxi1500-RawData \citep{ma-etal-2025-taxi1500}, as well as collections of poetry \citep{geocitiesYohanesManhitu} and pantun\footnote{A form of traditional Indonesian poetry} \citep{geocitiesPantunMelayu}. The complete statistics of the monolingual dataset used are presented in Table~\ref{tab:fasttext_dataset}.

\begin{table}[!h]
    \centering
    \resizebox{\linewidth}{!}{
    \begin{tabular}{lcccc}
    \toprule
        Dataset & \# of Texts & Avg. Words & Max. Words & Min. Words \\
    \midrule
        Tapaleuk News & 99 & 704.42 & 1,078 & 227 \\
        Jakarta Field Station & 66 & 6.24 & 13,184 & 879 \\
        Taxi1500-RawData & 18,452 & 43.31 & 284 & 1 \\
        Poem \& Pantun & 37 & 56.02 & 181 & 18 \\
    \bottomrule
    \end{tabular}
    }
\caption{Statistics of monolingual dataset in Kupang Malay.}
\label{tab:fasttext_dataset}
\end{table}

\subsection{Backbone Models}
We utilize the Cendol mT5 model \citep{cahyawijaya-etal-2024-cendol}, an open-source multilingual LLM supporting Indonesian and 18 local languages. Specifically, three variants of the Cendol mT5 model are employed; ranging from 350 and 580 million parameters to 1.2 billion parameters, respectively. To maintain model’s generalization, the Experience Replay technique \citep{NEURIPS2019_fa7cdfad} is applied. We used 1{,}000 additional training samples from the original Cendol model’s training collection\footnote{\url{https://huggingface.co/datasets/indonlp/cendol_collection_v2}} and combined them with the constructed instruction data for training. In addition, due to computational and storage constraints, optimization strategies are applied. This including the use of Brain Floating Point 16-bit (BF16) for reduced precision computation, gradient checkpointing to store partial activations before the backward pass, and gradient accumulation over two steps. All training parameters are detailed in Table~\ref{tab:hyperparams}.

\begin{table}[!t]
    \centering
    \resizebox{\linewidth}{!}{
        \begin{tabular}{lccc}
        \toprule
        \textbf{Parameter} & \textbf{mT5 small} & \textbf{mT5 base} & \textbf{mT5 large} \\
        \midrule
        max\_input\_length & 128 & 128 & 128 \\
        max\_output\_length & 52 & 52 & 52 \\
        batch\_size & 32 & 16 & 16 \\
        bfp16 & True & True & True \\
        lr & 3e-5 & 3e-5 & 3e-5 \\
        \bottomrule
        \end{tabular}
    }
    \caption{Hyperparameter settings for Cendol mT5 models (small to large).}
    \label{tab:hyperparams}
\end{table}

\subsection{Computational Resources}
All experiments were conducted using a single 40GB A100 GPU for all model training variations, and a single 8GB GeForce RTX 4060 Ti for evaluation and additional inference resources. The training and evaluation processes took a total of approximately $\sim$13 days, including two days each for training the small, base, and large model sizes. Additionally, the Weights \& Biases platform was used to monitor model training and GPU utilization, providing visualization and model tracking capabilities.

\subsection{Evaluation Suite}
\paragraph{Model Baseline}
The model baseline consists of comparisons between models trained with standard instructions and those with linguistic instructional prompts. The models using standard instructions were trained using the Cendol-mT5 model in small, base, and large parameter sizes. To ensure comparability, all model training parameters were kept consistent (\S\ref{tab:hyperparams}) and following the instruction template for the baseline model, adopted from \citet{cahyawijaya-etal-2024-cendol} for translation tasks. 


\paragraph{Comparison with Other Models} We conduct evaluation using both zero-shot and few-shot prompting on multilingual LLMs.
Specifically, the models used include BLOOMZ-7B1-MT and mT0-XXL-MT \citep{muennighoff-etal-2023-crosslingual}, Sailor-7B-Chat \citep{dou-etal-2024-sailor}, Aya-Expanse-8B \citep{dang2024ayaexpansecombiningresearch}, SeaLLMs-v3-7B-Chat \citep{zhang2024seallms3openfoundation}, Cendol-LLaMA2-7b-Inst, and Cendol-mT5-XL-Inst \citep{cahyawijaya-etal-2024-cendol}. 
Furthermore, to the best of our knowledge, the only existing Kupang Malay translation models are the multilingual Madlad400-3B-MT and Madlad400-7B-MT models by Google \citep{kudugunta2023madlad400multilingualdocumentlevellarge}. These models were trained on 100 billion sentences across 419 languages, including 25.4 thousand Kupang Malay sentences, all sourced from Bible texts. Additionally, Madlad400 requires a language prefix \texttt{<2{trg}>} in the input sentence.

\paragraph{Human Evaluation}
We conduct further evaluation by human evaluators to maintain the quality of translations. More specifically, evaluators were asked to rate the translation outputs using two evaluation metrics: adequacy and fluency. Each metric follows a four-level scale and conducted by recruiting several native speakers to assess and rate the translation outputs. The rankings were analyzed using inter-annotator agreement, measured with the Kappa coefficient. 
    

\section{Result \& Analysis}

\subsection{Comparison with Baseline}
\begin{table}[!t]
    \centering
    \resizebox{\linewidth}{!}{
    \begin{tabular}{lcccc}
        \toprule
        \textbf{Model} & \textbf{SacreBLEU} & \textbf{chrF++} & \textbf{TER} & \textbf{ROUGE-L} \\
        \toprule
        \multicolumn{5}{c}{Standard Prompt} \\
        \midrule
        Cendol-mT5-small-tuned & 6.62 & 23.79 & 102.5 & 19.97 \\
        Cendol-mT5-base-tuned & 8.77 & 26.91 & 98.08 & 25.23  \\
        Cendol-mT5-large-tuned & \textbf{8.97} & \textbf{27.30} & \textbf{97.13} & \textbf{25.94} \\
        \midrule
        \multicolumn{5}{c}{Independent Prompt} \\
        \midrule
        Lius-Large-Contextual & \textbf{9.47} & \textbf{27.78} & 96.90 & 26.55 \\
        Lius-Large-Phonetic & 9.42 & 27.63 & 97.52 & 26.48 \\
        Lius-Large-List-Group-Label & 9.37 & 27.85 & \textbf{96.41} & \textbf{26.86} \\
        Lius-Large-Semantic & 9.21 & 27.70 & 96.83 & 26.55 \\
        \midrule
        \multicolumn{5}{c}{Instructional Linguistic Prompt} \\
        \midrule
        Lius-Small-MT & 8.72 & 26.70 & 97.96 & 24.99 \\
        Lius-Base-MT & 11.63 & 30.26 & 92.90 & 29.75  \\
        Lius-Large-MT & \textbf{12.99} & \textbf{32.01} & \textbf{90.36} & \textbf{31.55}\\
        \bottomrule
    \end{tabular}
    }
    \caption{Performance comparison between \texttt{Lius} and Baseline Models. Ours is denoted as \textbf{Instruction Linguistic Prompt}, while \textbf{Independent Prompt} is independent instructional linguistic variants.}
    \label{tab:main-results}
\end{table}
Overall, the model performance across all experiments is shown in Table~\ref{tab:main-results}. \texttt{Lius} models achieved superior performance across all evaluation metrics, with the 1.2 billion parameters model demonstrating the highest average performance by 2–6 points on each metric and among the \texttt{Lius} variants. Notably, even the smaller variant with 300 million parameters, the model outperformed the standard instruction model of the same size by two points. This indicates that our approach remains effective even for smaller models in low-resource scenarios. In addition, all independent instructional outperformed the standard instruction models. For instance, SacreBLEU scores for these variants ranged from 9.21 to 9.47, surpassing the best-performing standard model. This highlights the benefit of providing task-specific instructions over generic templates.

\begin{figure}[htbp]
    \centering
    \begin{subfigure}{0.3\linewidth}
        \centering
        \includegraphics[width=\linewidth, trim=0 0 0 0, clip]{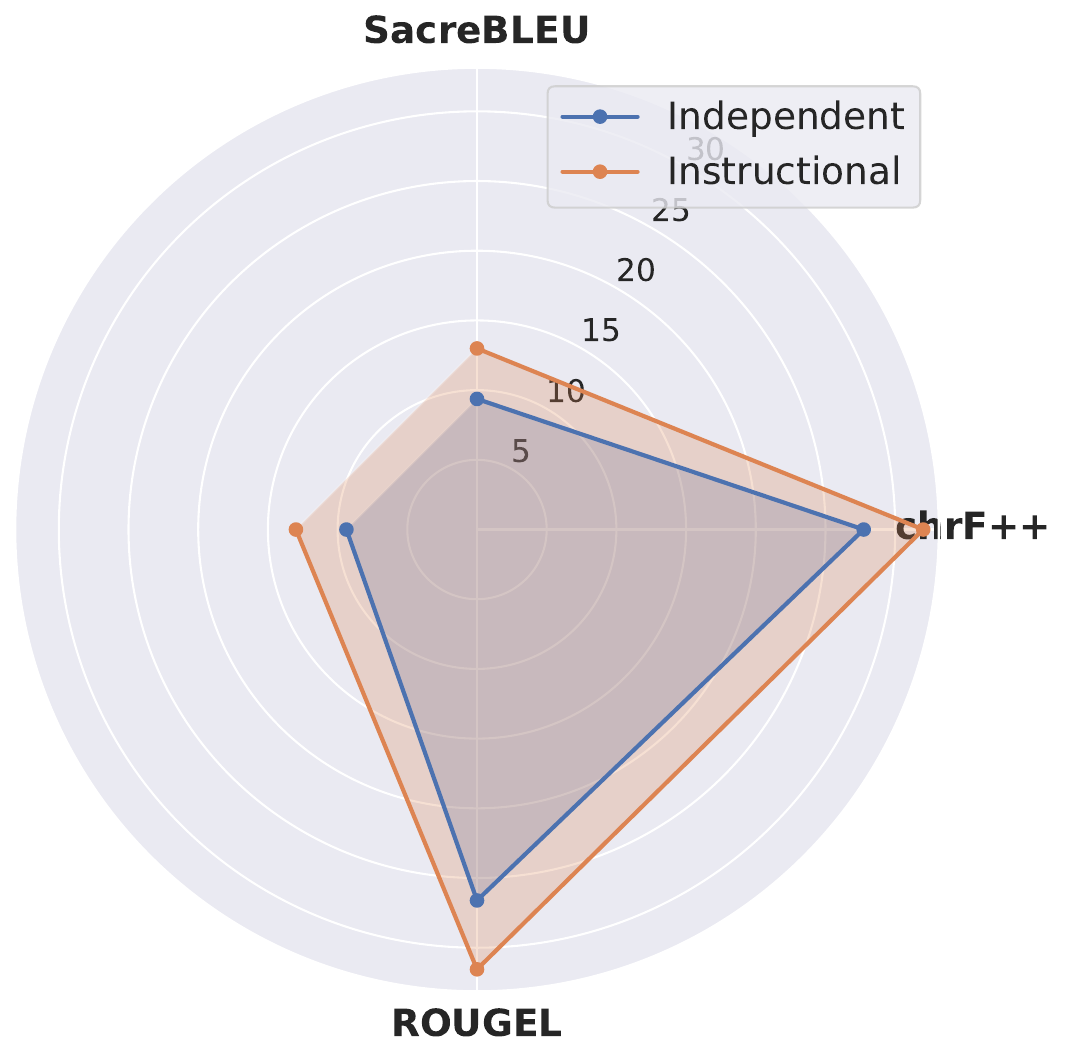}
        \caption{} 
        \label{fig:radar_instruksional}
    \end{subfigure}
    \hfill
    \begin{subfigure}{0.5\linewidth}
        \centering
        \includegraphics[width=\linewidth, trim=0 0 0 0, clip]{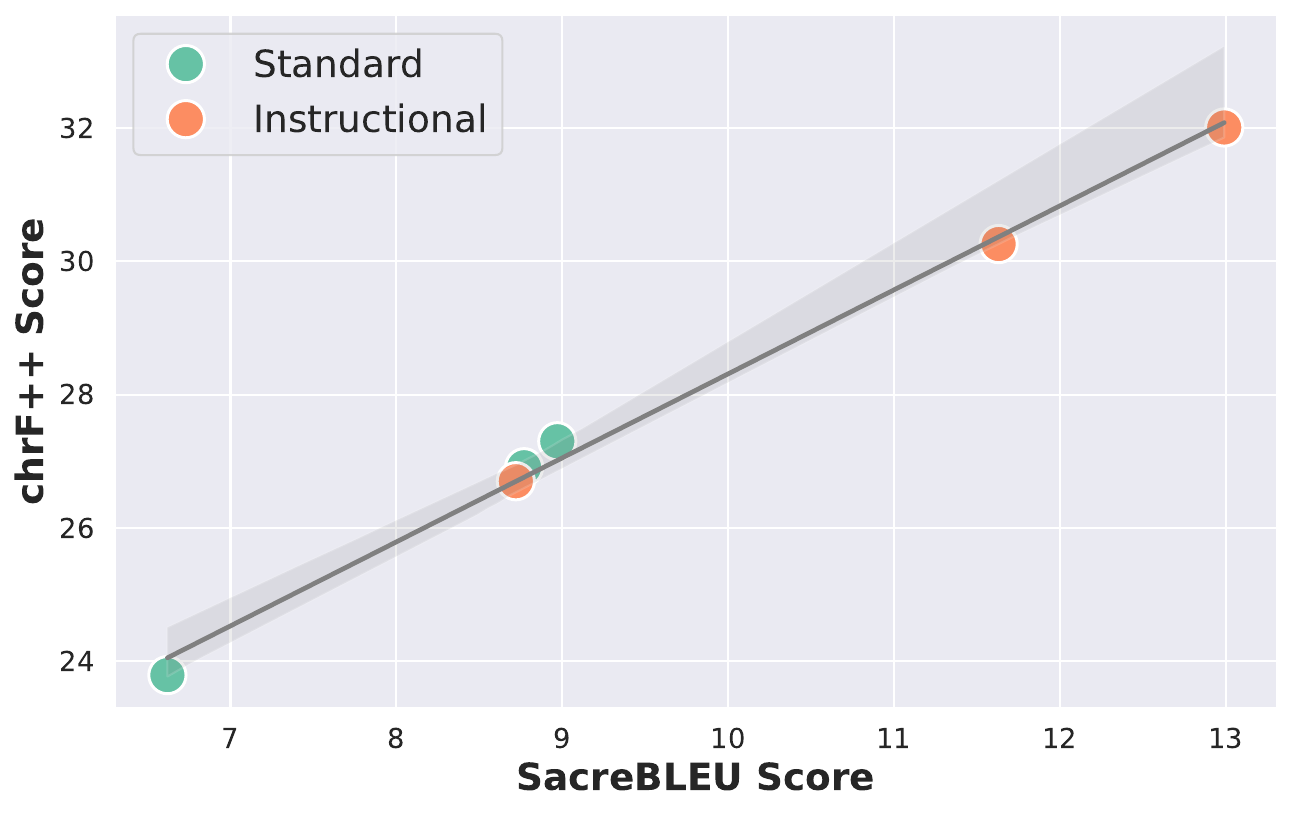}
        \caption{} 
        \label{fig:korelasi_bleu_chrf}
    \end{subfigure}
    \caption{(a) \texttt{Lius} performance and its independent instructional models using Large model. (b) Correlation between BLEU and chrF++ metrics.}
    \label{fig:instruction_comparison}
\end{figure}

\paragraph{Semantic and Contextual Analysis}
\texttt{Lius} improving model performance by approximately 4–6 points across all evaluation metrics. This indicates that our approach enables the model to capture deeper semantic features and comprehend sentence context. For instance, the lower score on the TER metric suggests that fewer edits are required to match the model’s translation output with the reference. This can be proved as denoted in Figure~\ref{fig:korelasi_bleu_chrf}, which shows the relationship between SacreBLEU and chrF++ scores, where models with higher BLEU scores also tend to have higher chrF++ scores. 


\paragraph{Independent Prompt Contribution}
We found that each independent instructional is slightly better than the standard baseline, which suggests that it contributes significantly to the performance of the \texttt{Lius} model. To better understand this finding, four example translations were selected and presented in Table~\ref{tab:independen-example}. There are different variations across the outputs from each instruction type. For example, in the target sentence \textit{"Botong sangka dia pung badan ba’isi, tagal su ampa bulan sonde datang bulan"}, the Semantic and Phonetic instructions signal the word \textit{"kotong"} (we), which semantically aligns with \textit{"botong"} (we). Furthermore, \texttt{Lius} is also capable of capturing the semantic meaning of the sentence. For instance, the phrase \textit{"balakang"} (restroom) is consistently translated as \textit{"kamar kici"} (small room/restroom) in Kupang Malay. 

\begin{table*}[!t]
    \centering
    \resizebox{\linewidth}{!}{
    \begin{tabular}{p{0.2\linewidth}p{0.2\linewidth}p{0.2\linewidth}p{0.2\linewidth}p{0.2\linewidth}p{0.2\linewidth}}
    \hline
        \textbf{Target} & \textbf{Context} & \textbf{Semantic} & \textbf{Phonetic} & \textbf{List-Group-Label} & \textbf{Instructional}   \\
    \hline
        Adu, be pung paru saki! Be parmisi pi balakang do. &
        aduh! beta pung perut saket. Mamisi, beta mau pi kamar kecil dolo. &  
        aduh! beta pung perut saket. Mamisi, beta mau pi kamar kecil dolo. &  
        adu! beta pung perut saket. Mamisi, beta mau pi kamar kecil dolo. &  
        aduh! beta pung perut saket banget. Maaf, beta mau pi kamar kecil dolo.
        &  
        Aduuu! Beta pung paru talalu saki. Tolong, beta pi kamar kici dolo. \\
    \hline
        Botong sangka dia pung badan ba'isi, tagal su ampa bulan sonde datang bulan. &
        Beta pikir dia hamil, karna su empat bulan sonde haid.& 
        kotong kira dia hamil, karna su empat bulan dia son haid. & 
        kotong kira dia hamil, karna su empat bulan dia tida haid. &
        Beta pikir dia hamil, karna su ampa bulan sonde haid. 
        &  
        Kotong kira dia hamil, te su ampat bulan son haid. \\
    \hline
    \end{tabular}
    }
\caption{Example of translation result across different instruction. }
\label{tab:independen-example}
\end{table*}

\paragraph{Training Process}
The training results of each model are illustrated in Figure~\ref{fig:model_comparison}. In general, models exhibit a steady downward trend in loss, indicating that the models are effectively learning and improving their translations over time. \texttt{Lius} with large parameter size began with a higher initial loss, however, as training progressed, it was able to significantly reduce its loss. In terms of convergence, the Small and Base models tended to converge faster than the Large model, primarily due to having fewer parameters to adjust. In addition, the computational cost and training efficiency of the Large model requires significantly more storage capacity as well as increased infrastructure during training. For instance, training duration increases exponentially, from 1 hour and 9 minutes to over 6 hours, reflecting a rise in computational complexity. 

\begin{table}[!h]
    \centering
    \caption{Efficiency Metrics of Lius Model Variants}
        \resizebox{\linewidth}{!}{
        \begin{tabular}{lcccc}
            \toprule
            Aspect & Lius-Small-MT & Lius-Base-MT & Lius-Large-MT \\
            \toprule
            Parameters & 350M & 580M & 1.2B \\
            Storage (GB) & 1.13 & 1.95 & 3.4 \\
            Training Time (hh:mm:ss) & 01:09:00 & 03:22:00 & 06:35:00 \\
            Memory Usage (\%) & 28 & 39 & 78 \\
            Inference Time (s) & 14.21 & 22.24 & 72.35 \\
            \bottomrule
        \end{tabular}
        }
    \label{tab:aspek-efisiensi}
\end{table}

\paragraph{Copying Behaviour}
Furthermore, we investigated a model’s tendency to directly copy tokens from the input sentence or Copying behaviour, which can degrades translation quality \citep{liu-etal-2021-copying}. We compute both copy accuracy
and copy rate
for each model, a higher copy value indicates a stronger tendency to replicate input tokens. We inspected that, both metrics exhibit a downward trend as model size increases. For instance, \texttt{Lius} with small parameter size shows the highest copy accuracy at 23.62\%, while the large one decreases to 20\%. Similarly, the Copy Rate decreases from 23.54\% in the Small model to 19.38\% in the Large model. 

\begin{figure}
    \resizebox{\linewidth}{!}{
        \begin{minipage}{0.45\linewidth}
            \resizebox{\linewidth}{!}{
                \includegraphics[trim=0 0 0 0,clip]{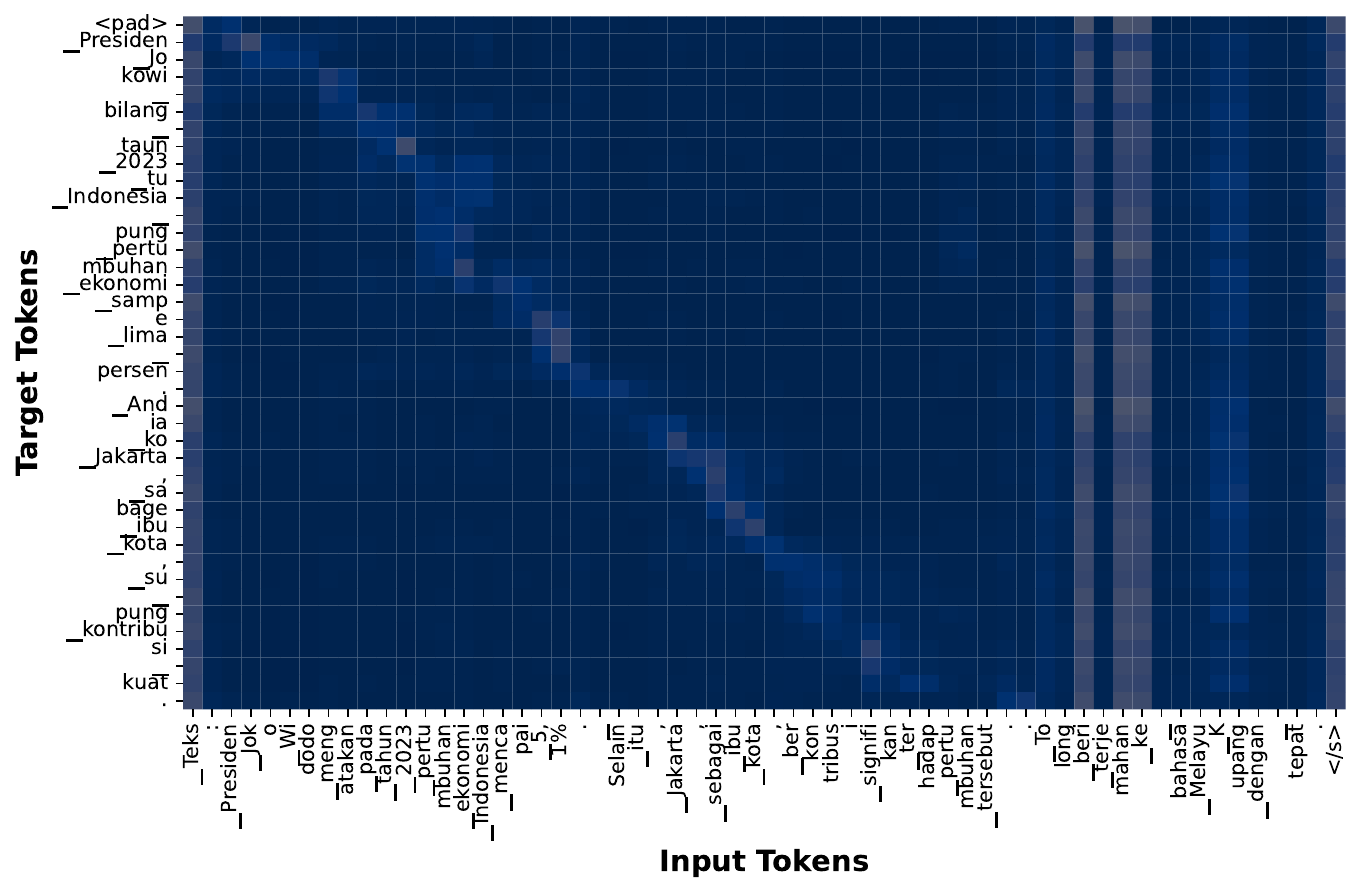}
            }
        \end{minipage}
        \hspace{3pt}
        \begin{minipage}{0.55\linewidth}
            \resizebox{\linewidth}{!}{
                \includegraphics[trim=0 0 0 0,clip]{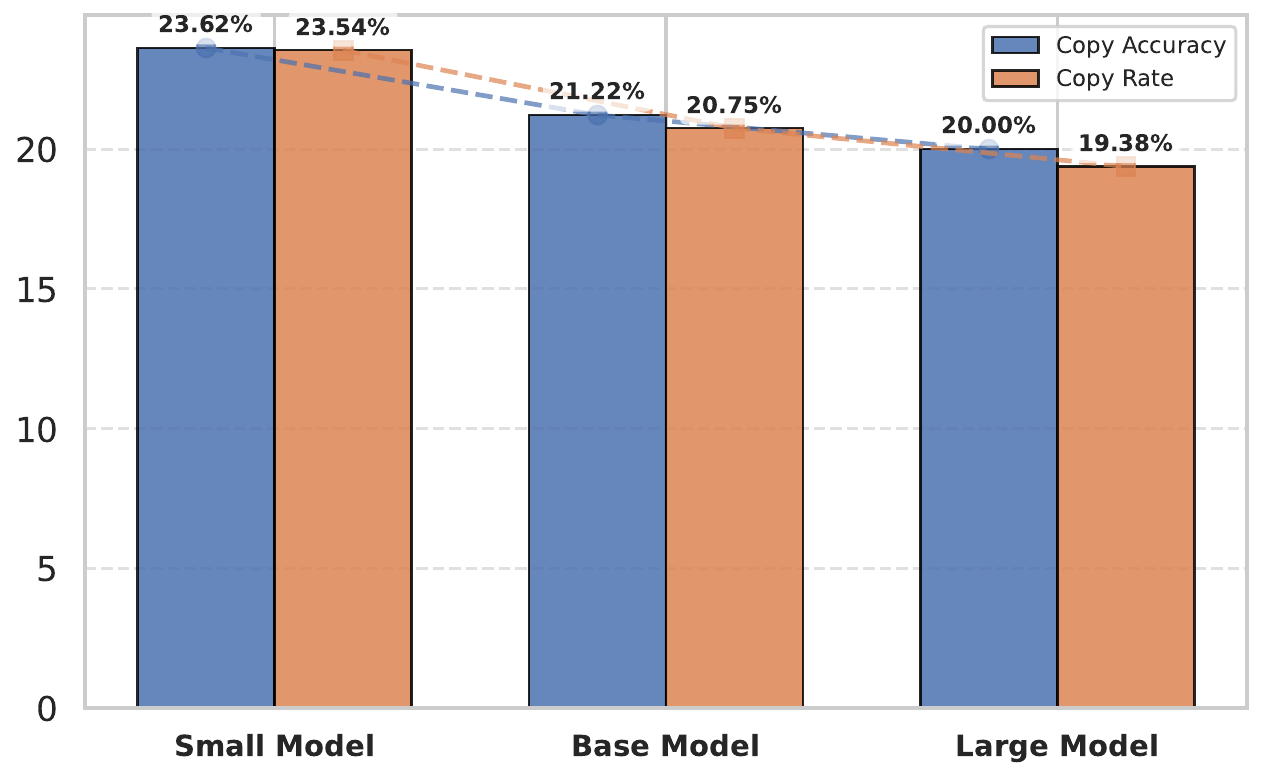}
            }
        \end{minipage}
    }
    \caption{Analysis of Copying Behaviour across Models}
    \label{fig:copying_behaviour}
\end{figure}

Furthermore, we found that the declining trend in both Copy Accuracy and Copy Rate in the Base and Large models implies that increased model capacity enables better comprehension of linguistic context, encouraging the model to translate rather than copy. For instance, as shown in Figure~\ref{fig:copying_behaviour}, words like "Presiden", "Jokowi", "2023", "Indonesia", and "Jakarta" are accurately copied by the model. Meanwhile, terms such as "tahun" are translated into "\textit{taon}" and "sebagai" into "\textit{sabage}", indicating direct translation by the model.

\subsection{Existing NMT \& LLMs Comparison}
\begin{table*}[!htbp]
    \centering
    \resizebox{\linewidth}{!}{
    \begin{tabular}{lcccccccc}
        \toprule
        \multirow{2}{*}{\textbf{Model}} &
        \multicolumn{4}{c}{\textbf{Zero-shot}} &
        \multicolumn{4}{c}{\textbf{Few-shot}} \\
        \cmidrule(lr){2-5} \cmidrule(lr){6-9}
        & \textbf{SacreBLEU} & \textbf{chrF}++ & \textbf{TER} & \textbf{ROUGEL} & \textbf{SacreBLEU} & \textbf{chrF}++ & \textbf{TER} & \textbf{ROUGEL} \\
        \toprule
        \multicolumn{9}{c}{Multilingual \& Regional LLMs} \\
        \midrule
        BLOOMZ-7B1-MT & 3.14 & 15.03 & 148.08 & 6.37 & 1.13 & 13.17 & 372.48 & 4.69 \\
        mT0-XL & 4.25 & 21.00 & 130.35 & 12.26 & 2.25 & 19.19 & 163.03 & 12.07 \\
        mT0-XXL-MT & 4.86 & 22.93 & 114.60 & 14.40 & 2.15 & 18.31 & 173.29 & 11.11 \\
        Sailor-7B-Chat & 3.47 & 22.36 & 154.16 & 13.81 & 3.47 & 22.36 & 154.16 & 13.81 \\
        Aya-Expanse-8B & 3.35 & 23.81 & 178.34 & 16.10 & -- & -- & -- & -- \\
        SeaLLMs-v3-7B-Chat & 4.44 & 21.80 & 146.95 & 15.95 & 1.12 & 10.88 & 294.83 & 4.46 \\
        Cendol-LLaMA2-7b-inst & 1.87 & 11.98 & 122.50 & 4.31 & 1.45 & 5.96 & 134.57 & 2.23 \\
        Cendol-mT5-XL-Inst & 3.72 & 20.68 & 143.90 & 11.00 & -- & -- & -- & -- \\
        \midrule
        \multicolumn{9}{c}{Multilingual NMT} \\
        \midrule
        Madlad400-3B-MT & 2.22 & 20.96 & 136.81 & 13.94 & -- & -- & -- & -- \\
        Madlad400-7B-MT & 2.22 & 20.36 & 128.79 & 13.32 & -- & -- & -- & -- \\
        \midrule
        \multicolumn{9}{c}{Lius Models} \\
        \midrule
        Lius-Small-MT & 9.20 & 27.75 & 98.76 & 25.55 & 8.72 & 26.70 & 97.96 & 24.99 \\
        Lius-Base-MT & 11.88 & 31.23 & 92.83 & 30.52 & 11.63 & 30.26 & 92.90 & 29.75 \\
        Lius-Large-MT & \textbf{13.27} & \textbf{32.80} & \textbf{89.64} & \textbf{32.62} & \textbf{12.99} & \textbf{32.01} & \textbf{90.36} & \textbf{31.55} \\
        \bottomrule
    \end{tabular}
    }
    \caption{Performance comparison of multilingual, regional LLM and NMT, and \texttt{Lius} models on zero-shot and few-shot prompting settings. There are two models excluded in Few-shot prompting due to limited computational time.}
    \label{tab:combined-results}
\end{table*}

We conduct comparison using Madlad400-3B-MT and Madlad400-7B-MT to measure the further capabilities of \texttt{Lius}. As shown in Table~\ref{tab:combined-results}, the multilingual NMT models achieve substantially lower scores across all evaluation metrics, whereas the \texttt{Lius} model demonstrates up to a sixfold improvement. This implies that although Madlad400 has a larger number of parameters its performance in translating Kupang Malay text remains highly limited.

Similarly, \texttt{Lius} model significantly outperforms regional LLMs across all evaluation metrics with significant margin on zero-shot and few-shot prompting. For instance, in few-shot scenario, multilingual models such as mT0-XXL-MT and BLOOMZ-7B1-MT relatively underperform, with SacreBLEU scores of only 2.15 and 1.13 respectively, and extremely high TER values (173.29 and 372.48). This indicates that internally, \texttt{Lius} has better semantic understanding, task awareness, and better alignment to translation tasks rather than the other multilingual and regional LLMs.

\subsection{Generalization Towards Unseen Data}
\begin{figure}[!h]
    \centering
    \includegraphics[width=0.97\linewidth,trim={0 0 0 0}, clip]{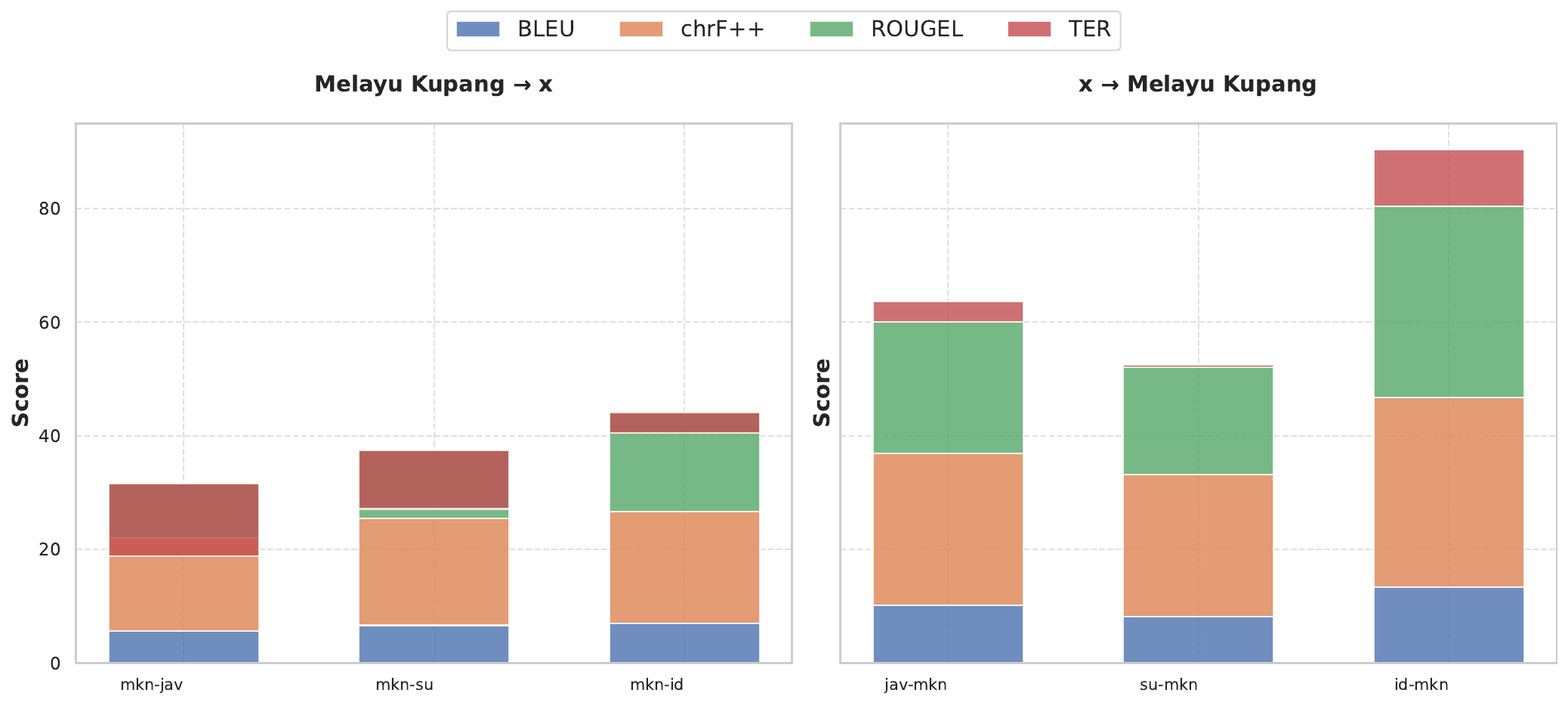}
    \caption{Bidirectional translation performance of the \texttt{Lius} Model.}
    \label{fig:heatmap-crosslingual}
\end{figure}

\paragraph{Unseen Language}
To evaluate multilingual capability of the model, we conduct an experiment involving Javanese, Sundanese, and Indonesian in a bidirectional translation setup using 100 randomly sampled sentences. The model's performance evaluation is shown in Figure \ref{fig:heatmap-crosslingual}. Surprisingly, the evaluation results indicate that the model retains strong multilingual signals. For example, translation from Javanese to Kupang Malay achieves SacreBLEU scores of 10.17, with chrF++ of 26.76 and ROUGE-L of 23.10. 
This result highlight the model's generalization on cross-lingual transfer.

\begin{table*}[!h]
    \centering
    \renewcommand{\arraystretch}{1.2}
    \setlength{\tabcolsep}{6pt}
    \resizebox{\linewidth}{!}{
    \begin{tabular}{p{0.3\linewidth} p{0.3\linewidth} p{0.3\linewidth}p{0.3\linewidth}}
    \toprule
        \textbf{Task} & \textbf{Input} & \textbf{Target} & \textbf{Output} \\
    \midrule
        Sentiment Analysis & 
        \textit{mung kamare mambu rokok lan banyu kamar mburi trocoh neng kamar karo AC kurang adem. Ngapurane komplen supoyo dijogo kualitase} & 
        negative  &
        negative \\  
    \midrule
        Question Answering - Does the text contain an element of surprise? & 
        \textit{simkuring nu isuk2 rek nyimpen pakean ka cucian jeung ninggali sirah orok bijil ti panto gudang (ieu sigana simkuring nganyimpang, soalna simkuring sieun pisan waktos eta)} &
        no &
        no \\ 
    \midrule
        Topic Modeling (\textit{fear}, \textit{disgusted}, \textit{sad}, \textit{happy}, \textit{angry}, \textit{surprise}, \textit{shame}) & 
        \textit{Alhamdulillah ammetai persib, rannui atia, tinroa tong sannang. MVP anne bangngia anjo [USERNAME] yang pa'kullei a'cini' persib na tena a'pau le'ba siallo susai a'komunikasi kah kasibukanganna ade bedur} &
        sad &
        sad \\ 
    \bottomrule
    \end{tabular}
    }
    \caption{Examples of multitasking performance on the \texttt{Lius} Model.}
    \label{tab:cross-task-example}
\end{table*}

\paragraph{Unseen Task}
In addition to its multilingual capability, we investigated the \texttt{Lius} multitasking ability. Based on the analysis presented in Table \ref{tab:cross-task-example}, the model demonstrates its multitasking capability despite being fine-tuned specifically for translation tasks.  For instance, in the sentiment analysis task, the model consistently classifies text with "negative" sentiment, while in the topic modeling task, the model accurately categorizes the input as "sad". These indicate the model's ability to capture cross-lingual emotional nuance in other languages, such as complex regional language and non-standard lexical variations. 

\subsection{Robustness Testing}
\begin{figure}[!h]
    \centering
    \includegraphics[width=0.8\linewidth,trim={0 0 0 0}, clip]{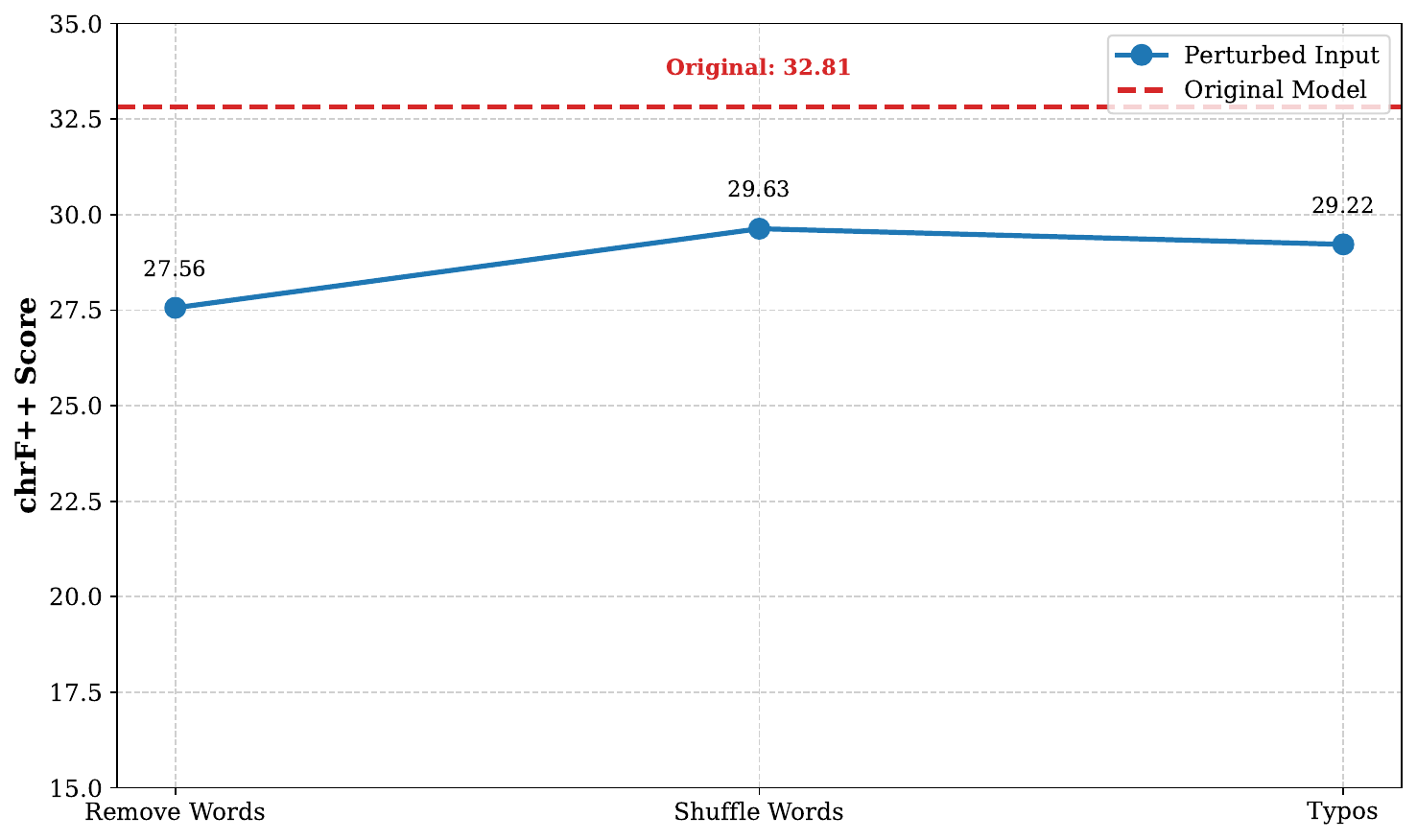}
    \caption{Comparison of model performance on perturbed inputs: Word Shuffling, Word deletion, and Typographical errors.}
    \label{fig:chrf++_compare}
\end{figure}

We conducted a series of evaluations on the \texttt{Lius} model to assess its robustness under various perturbations applied to the input text. Overall, as shown in Figure \ref{fig:chrf++_compare}, the model performs reasonably well on modified sentences, with the performance gap relatively small compared to the unmodified input. These results demonstrate the model’s ability to reconstruct structured and contextually appropriate sentences despite input variations. 

\subsection{Human Evaluation}
\begin{figure}[htbp]
    \centering
    \begin{subfigure}{0.45\linewidth}
        \centering
        \includegraphics[width=\linewidth, trim=0 0 0 0, clip]{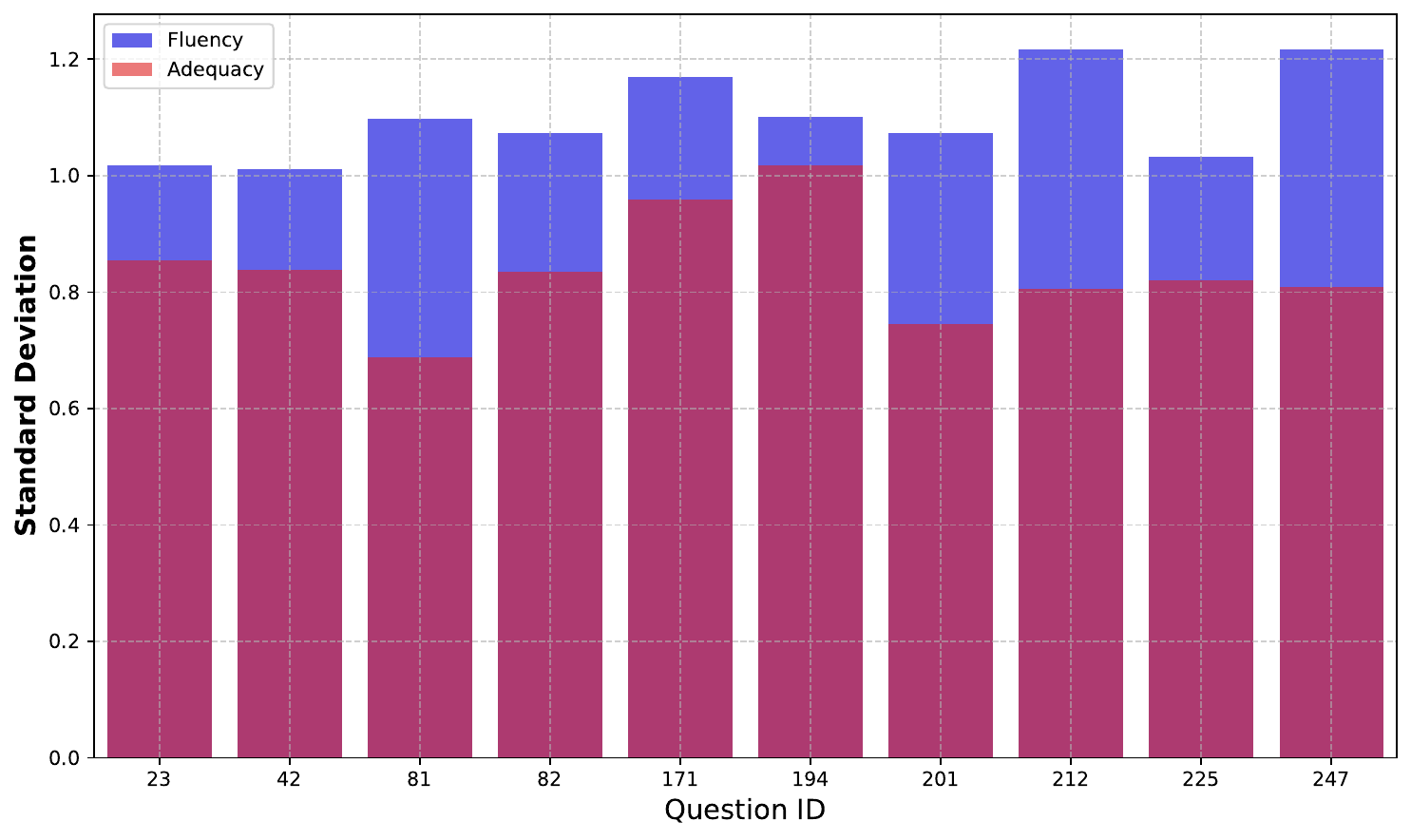}
        \caption{} 
        \label{fig:variabilitas_fluency_adequacy}
    \end{subfigure}
    \hfill
    \begin{subfigure}{0.45\linewidth}
        \centering
        \includegraphics[width=\linewidth, trim=0 0 0 0, clip]{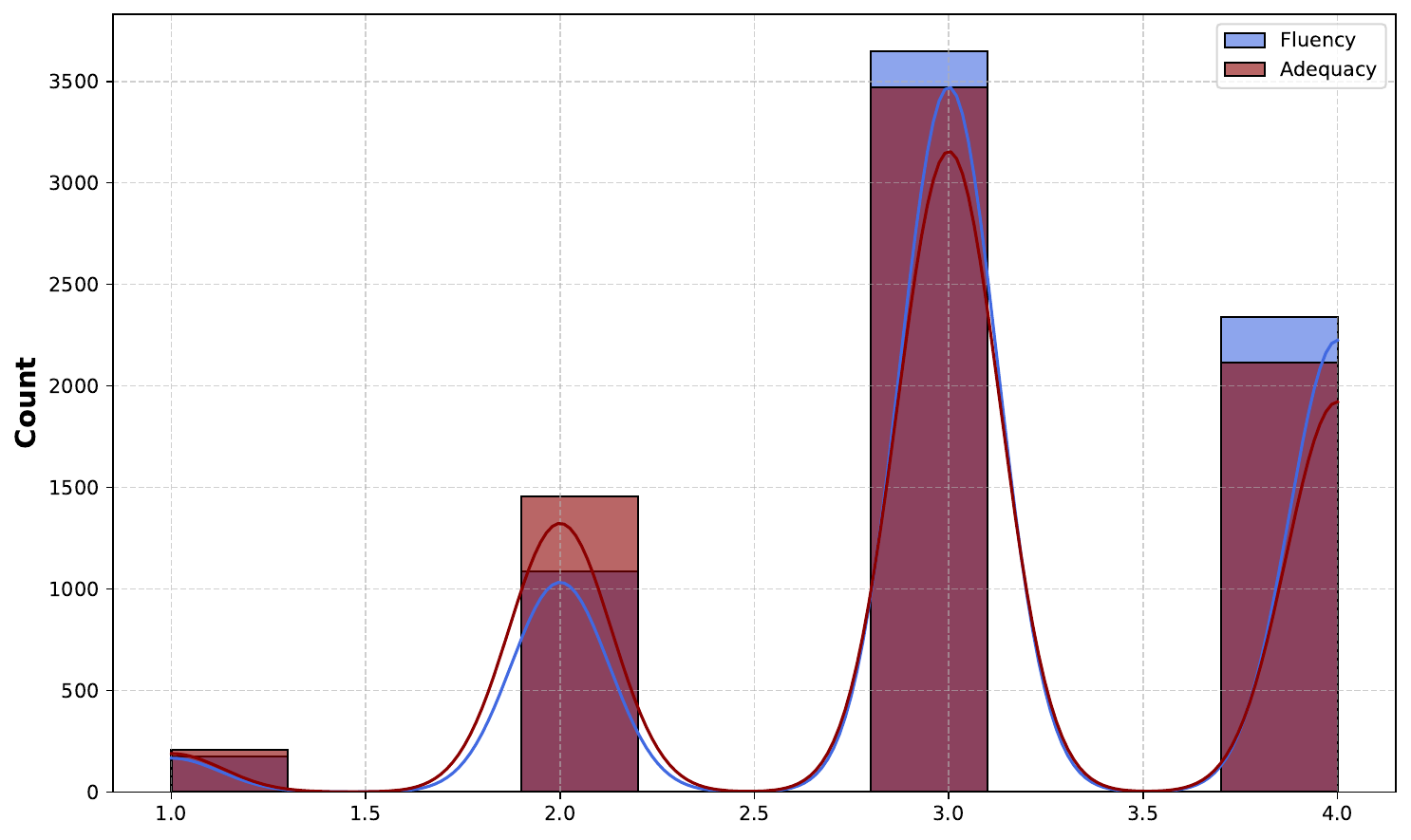}
        \caption{} 
        \label{fig:distribusi_fluency_adequcy}
    \end{subfigure}
    \caption{(a) Variability score for fluency and adequacy. (b) Distribution score for translation quality based on fluency and adequacy.}
    \label{fig:human_metrik}
\end{figure}

\paragraph{Quality Rating Distribution}
Human Evaluation is conducted by involving native speakers to assess the translation results of the model. In general, the average translation scores fall within the range of 3 to 4, presented in Figure \ref{fig:distribusi_fluency_adequcy}, indicating that the majority of the translation outputs are considered reasonably good in terms of both fluency and adequacy. However, the adequacy score distribution appears slightly more dispersed, suggesting that conveying meaning accurately in some translations remains a challenge. Furthermore, some evaluators tend to consistently assign higher or lower score, may reflect subjectivity, and individual preferences. On the other hand, there is greater disagreement among raters in assessing sentence naturalness, as shown in Figures \ref{fig:variabilitas_fluency_adequacy}, which the standard deviation of fluency ratings is generally higher than that of adequacy.

\paragraph{Inter-annotator Agreement}
In addition to analyzing score distribution and inter-rater variability, the ordinal \textit{Cohen’s Kappa} coefficient was also calculated to evaluate the level of agreement among evaluators in scoring each translation output. To mitigate potential bias from evaluators who consistently assign overly high or low scores, score normalization was applied prior to the calculation. The results indicate that the weighted Kappa value for the adequacy  is 0.2099, while for fluency it is 0.1339. According to the interpretation scale by \citet{264269a5-5657-3fa1-9d92-ddad5e679245}, both values fall within the slight agreement category, suggesting a low level of agreement among evaluators. We suspected that, internally, Kupang Malay does not yet have a widely recognized standardized guideline, even though it has evolved into a distinct language \citep{jacob2003kamus}. This lack of standardization allows for divergent perceptions among speakers when evaluating translation quality. 

\section{Conclusion}
This study proposes a method called the linguistic instructional approach to train LLMs based on the Continual Instruction Tuning paradigm. It incorporates specific instructions and enables iterative instruction-based during training process. The model is trained using the Cendol-mT5 architecture and demonstrates superior performance compared to models trained with standard instructions, outperforming both NMT models and multilingual LLMs in zero-shot and few-shot prompting scenarios, with performance gaps ranging from 10 to 13 points across various metrics. The model also retains its multilingual capabilities and performs competently on other tasks such as sentiment analysis and question answering. A human evaluation was also conducted, showing that the model still fall short of high-quality standards as judged by native speakers. The lack of a standardized guideline for Kupang Malay poses a considerable challenge in the human evaluation process.


\bibliography{tacl2021}
\bibliographystyle{acl_natbib}

\newpage

\appendix

\section{Robustness Testing Examples}
\label{app:robustness_testing}
\paragraph{Typographical errors} As shown in Table \ref{tab:kesalahan_ketik}, words such as "dan" are altered to "daan", "bajunya" to "bjunya", and "kemudian" to "emudian". Nevertheless, the model is capable of capturing the intended meaning from such noisy input. In some cases, the model successfully maps distorted words to semantically relevant expressions, such as transforming "emudian" into \textit{"abis itu"} in Kupang Malay. This demonstrates the model’s capacity to understand and reconstruct meaning despite distortions introduced by typographical noise.

\begin{table}[!h]
    \centering
    \resizebox{\linewidth}{!}{
    \begin{tabular}{p{0.3\linewidth} p{0.3\linewidth} p{0.3\linewidth}}
    \toprule
        \textbf{Original Text} & \textbf{Typos} & \textbf{Output} \\
    \midrule
         Setelah itu, baptua melipat koran dan kemudian menyimpannya di dalam bajunya kemudian pulang. & 
         Setelah itu, baptua melipat koran daan kemudian menyimpannya di dalam bjunya emudian pulang. &
         \textit{Abis itu, baptua lipat surat ko simpan di dia pung buk} abis itu pulang.\\ 
    \bottomrule
    \end{tabular}
    }
\caption{Examples of Translations with Typographical Errors}
\label{tab:kesalahan_ketik}
\end{table}

\paragraph{Word deletion} As shown in Table \ref{tab:penghapusan_kata}, the phrase "kau akan mendapatkan tinjuku" is reduced to "kau tinjuku", which drastically alters the intended meaning. The output generated after word deletion reveals that the model struggles with such distortions, occasionally inserting unrelated elements such as \textit{“bini”} and \textit{“matua”}. This illustrates the model's limitations in grasping holistic meaning when faced with incomplete input, leading to outputs that are not only inaccurate but also enriched with hallucinated content.

\begin{table}[!h]
    \centering
    \resizebox{\linewidth}{!}{
    \begin{tabular}{p{0.3\linewidth} p{0.3\linewidth} p{0.3\linewidth}}
    \toprule
        \textbf{Original Text} & \textbf{Deletion} & \textbf{Output} \\
    \midrule
         Biarkan dia pergi atau kau akan mendapatkan tinjuku, Ama Jola sudah ingin melompat pada orang tua. & 
         Biarkan dia pergi atau kau tinjuku, Ama Jola ingin melompat pada tua. &
         \textit{Biar itu su jalan ko lu puku be pung bini, Ama Jola mo malompa sang matua} \\ 
    \bottomrule
    \end{tabular}
    }
\caption{Examples of Translations with Word Deletion}
\label{tab:penghapusan_kata}
\end{table}

\paragraph{Word Shuffling} As shown in Table \ref{tab:pengacakan-kalimat}, where the word "pada" is randomly misplaced. Nevertheless, the model demonstrates the ability to reconstruct structured and contextually appropriate sentences. Although the translation of the word \textit{"bagitu"} is inaccurate, the model still manages to produce diverse outputs that reflect resilience to structural perturbations in the input.

\begin{table}[!h]
    \centering
    \renewcommand{\arraystretch}{1.2}
    \setlength{\tabcolsep}{6pt}
    \resizebox{\linewidth}{!}{
    \begin{tabular}{p{0.3\linewidth} p{0.3\linewidth} p{0.3\linewidth}}
    \toprule
        \textbf{Original Text} & \textbf{Shuffled} & \textbf{Output} \\
    \midrule
         tapi aku tidak berpikir untuk menikah pada saat itu. & 
         tapi aku tidak berpikir pada menikah untuk saat itu. &
         \textit{tapi itu waktu be son berpikir untuk menika bagitu}.\\ 
    \bottomrule
    \end{tabular}
    }
\caption{Examples of Translations with Shuffled Input}
\label{tab:pengacakan-kalimat}
\end{table}

\section{Human Evaluation Process}
\label{app:human_evaluation}
\begin{figure}[!h]
    \includegraphics[width=\linewidth,trim={0, 0, 0, 0}, clip]{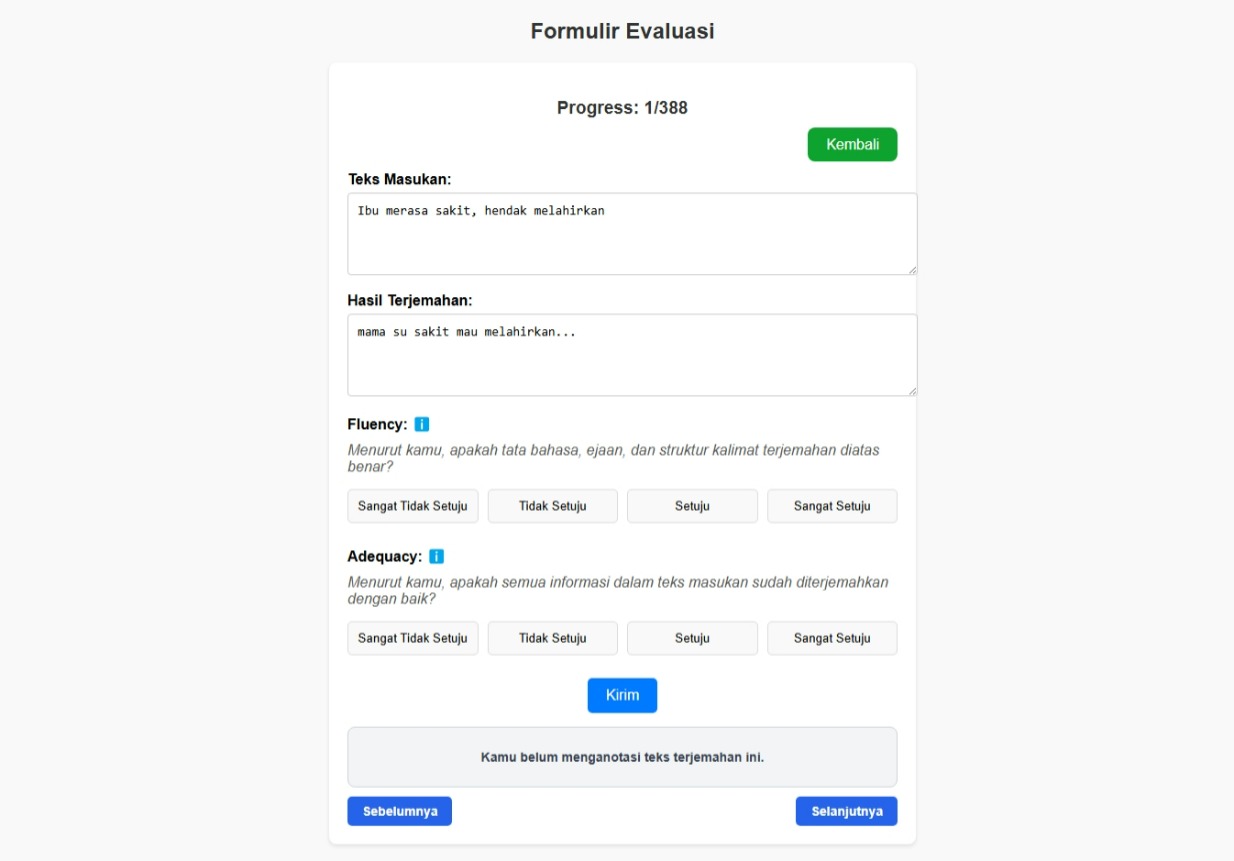}
    \caption{User Interface of Annotation Application}
    \label{fig:formulir}
\end{figure}

Human evaluation was conducted by involving native speakers to assess the translations produced by the \texttt{Lius} model. We evaluated based on two main questions: \textbf{In your opinion, is the grammar, spelling, and sentence structure of the translation correct?}—which corresponds to the \textit{fluency} metric; and \textbf{In your opinion, has all the information in the source text been accurately translated?}—which corresponds to the \textit{adequacy} metric. The evaluation took place over approximately two months, from January 21 to March 21, 2025. The process was carried out through a web-based annotation platform built using the Python programming language\footnote{\url{https://www.python.org/}} and the Flask framework\footnote{\url{https://flask.palletsprojects.com/en/stable/}}. The user interface of the annotation platform is shown in Figure~\ref{fig:formulir}. 

\begin{table}[!h]
    \centering
    \caption{Evaluation Scale for Adequacy and Fluency}
    \resizebox{\linewidth}{!}{
    \begin{tabular}{ccc}
        \toprule
        \textbf{Score} & \textbf{Adequacy} & \textbf{Fluency} \\
        \midrule
        4 & All meaning preserved & Very fluent \\
        3 & Most meaning preserved & Fluent \\
        2 & Some meaning preserved & Moderately fluent \\
        1 & Little meaning preserved & Not fluent \\
        \bottomrule
    \end{tabular}
    }
    \label{tab:skala-evaluasi}
\end{table}

The evaluators came from diverse backgrounds, including university students, lecturers, high school teachers, civil servants, and government language development officials. They were mostly residents of Kupang Malay-speaking areas, such as the cities of Kupang, SoE, and parts of Sabu Island. Of the 40 respondents registered in the system, 18 successfully completed the full evaluation. Four evaluators were excluded due to providing identical scores across all translation samples. Examples of the evaluation results for input, translation, and their corresponding fluency and adequacy scores are presented in Table~\ref{tab:data-input-terjemahan}.

\begin{table*}[!t]
    \centering
    \resizebox{\linewidth}{!}{
    \begin{tabular}{c p{6cm} p{6cm} c c}
        \toprule
        \textbf{ID} & \textbf{Input Text} & \textbf{Translation Output} & \textbf{Fluency} & \textbf{Adequacy} \\
        \midrule
        4986 & Tolong kalian pikirkan sendiri. Yang mana yang lebih baik? Kami mengikuti perintah bapak-bapak, atau kami mengikuti perintah Tuhan Allah? & Pikir sa. Mana yang lebe bae? Kotong iko bapa-bapa pung parenta, ko kotong iko Tuhan pung parenta? & 3 & 2 \\
        4987 & ekor kuda & kuda pung ekor & 4 & 4 \\
        4988 & Saya belum begitu paham dengan maksudnya. Bisakah engkau menerangkannya kembali kepada saya? & Beta balóm mangarti deng dia pung maksud bagini. Bosong bisa carita lai sang beta? & 2 & 3 \\
        4989 & Anak kecil harus tidur lebih awal. & ana kici harus tidor dolo. & 2 & 4 \\
        4990 & Ibu merasa sakit, hendak melahirkan & mama su sakit mau melahirkan... & 3 & 3 \\
        4991 & He! Jalanlah cepat! Jangan lambat seperti pengantin. Nanti kita ketinggalan kendaraan. Akhirnya kita hanya duduk termangu-mangu saja. & He! Jalan capat! Jang lambat sama ke pengantin, nanti kotong katinggal kendaraan, akirnya kotong dudu ragu-ragu sa. & 3 & 2 \\
        4992 & Waktu dia baru tiba dari Jakarta, dia pergi ke Kolhua untuk menegur-sapa pamannya & baru datang dari Jakarta dia pi Kolhua untuk tegur dia pung om. & 3 & 2 \\
        4993 & Jangan duduk mengangkang. Nanti anak laki-laki bisa melihat celana dalammu. & Jang dudu mangangkang, nanti ana laki laki liat lu pung calana dalam. & 4 & 2 \\
        4995 & Marilah kita bermain sembunyi, yuk? Engkau yang mencari dan aku yang bersembunyi, ya? & Mari kotong maen sembunyi, a? Lu yang cari, beta yang sambunyi, a? & 4 & 2 \\
        \bottomrule
    \end{tabular}
    }
    \caption{Examples of Fluency and Adequacy Input, Translation, and Evaluation Data}
    \label{tab:data-input-terjemahan}
\end{table*}

\section{Instruction Template}
The instruction template refers to direct commands used for translation tasks and constitutes a core component of the prompt template, marked with the label \textbf{INSTRUCTION}. A total of nine types of instruction template variations were developed, including styles such as formal-direct commands, direct questions, and narrative-style prompts. Each variation consists of 3 to 5 instructions, resulting in a total of 50 distinct instructions. Examples of these instruction template variations are shown in Table~\ref{tab:template_instruksi}. The \textbf{INPUT} represents the sentence to be translated, \textbf{SOURCE} denotes the source language (Indonesian), and \textbf{TARGET} refers to the target language (English). For each data instance, the selection of an instruction template is performed randomly.

\begin{table}[!h]
    \centering
    \resizebox{\linewidth}{!}{
    \begin{tabular}{lp{0.75\linewidth}}
        \toprule
        \textbf{Variasi} & \textbf{Prompt} \\
        \midrule
        Formal \& Langsung & Teks dalam bahasa \{\textbf{SOURCE}\}: \{\textbf{INPUT}\}\textbackslash nTerjemahkan ke dalam bahasa \{\textbf{TARGET}\}: \\        
        Pertanyaan & Apa terjemahan dari teks ini dalam bahasa \{\textbf{TARGET}\}?\textbackslash nTeks: \{\textbf{INPUT}\} \\
        Konteks & \{\textbf{INPUT}\} adalah teks dalam bahasa \{\textbf{SOURCE}\}.\textbackslash nSilakan terjemahkan ke bahasa \{\textbf{TARGET}\}. \\
        Spesifik & Mohon bantu menerjemahkan teks berikut dari bahasa \{\textbf{SOURCE}\} ke bahasa \{\textbf{TARGET}\}.\textbackslash nTeks: \{\textbf{INPUT}\} \\
        Konteks Pengguna & Jika Anda membaca teks berikut dalam bahasa \{\textbf{SOURCE}\}: \{\textbf{INPUT}\} \textbackslash nBagaimana Anda akan menyampaikannya dalam bahasa \{\textbf{TARGET}\}? \\
        \bottomrule
    \end{tabular}
    }
    \caption{5 Examples of Instruction Template Variations}
    \label{tab:template_instruksi}
\end{table}

Examples of single input data entries for each type of instruction can be seen in Table~\ref{tab:contoh_prompt_list_group_label}, Table~\ref{tab:contoh_prompt_context}, Table~\ref{tab:contoh_prompt_synonym}, and Table~\ref{tab:contoh_prompt_makna}. For instance, in the case of the List-Group-Label-based instruction, the sentence {\color{violet}\texttt{Aku makan kue tiga potong.}} serves as the input text, while sentences such as {\color{purple}\texttt{Label 1: kala, siri, cop}} and {\color{purple}\texttt{Label 3: fiik, binatang, babika}} represent the List-Group-Label-type instruction. The sentence {\color{olive}\texttt{Berikut adalah kategori kata dalam bahasa Melayu Kupang:}} is the prompt template, and {\color{teal}\texttt{Teks berikut membutuhkan terjemahan dari bahasa Indonesia ke bahasa Melayu Kupang: \texttt{Terjemahan:}}} serves as the instruction template.

\begin{table*}[ht!]
    \setlength{\tabcolsep}{6pt}
    \centering
    \footnotesize
    \resizebox{\linewidth}{!}{
    \begin{tabular}{r p{0.84\linewidth}}
    \toprule
    \textbf{Input:} & \color{olive} \texttt{Berikut adalah kategori kata dalam bahasa Melayu Kupang:} \\[-0.8ex]
    & \color{purple}\texttt{Label 1: kala, siri, cop} \\[-0.8ex]
    & \color{purple}\texttt{Label 2: saprei, Mabok sonde jadi-jadi, alas paru} \\[-0.8ex]
    & \color{purple}\texttt{Label 3: fiik, binatang, babika} \\[-0.8ex]
    & \color{purple}\texttt{Label 4: tiga-tiga, tigapulutiga, tiga} \\[-0.8ex]
    & \color{purple}\texttt{Label 5: bet, rabe, be} \\[-0.2ex]
    & \color{teal}\texttt{Teks berikut membutuhkan terjemahan dari bahasa Indonesia ke bahasa Melayu Kupang:} \color{violet}\texttt{Aku makan kue tiga potong.} \\[-0.1ex]
    & \color{teal}\texttt{Terjemahan: } \\[-0.1ex]
    
    \textbf{Output:} & \texttt{Beta makan kue tiga potong.} \\
    \bottomrule
    \end{tabular}
    }
    \caption{Example of List-Group-Label Based Instruction Data}
    \label{tab:contoh_prompt_list_group_label}

\end{table*}

\begin{table*}[ht!]
    \setlength{\tabcolsep}{6pt}
    \centering
    \footnotesize
    \resizebox{\linewidth}{!}{
    \begin{tabular}{r p{0.84\linewidth}} 
    \toprule
    \textbf{Input:} & \color{olive} \texttt{Anda memiliki kalimat berikut dalam bahasa Melayu Kupang:} \\[-0.8ex] 
    & \color{purple}\texttt{- Beta tar paksa lu ko iko sang beta. Ma kalo terpaksa, na iko sa} \\[-0.8ex] 
    & \color{purple}\texttt{- Dia rasa tar enak dalam hati, tagal dia pung ana pung jahat} \\[-0.1ex] 
    & \color{teal}\texttt{Terjemahkan teks berikut dari bahasa Indonesia ke bahasa Melayu Kupang.} \\[-0.2ex] 
    & \color{teal}\texttt{Teks: }\color{violet}\texttt{Saya juga, untuk pertama kalinya (saya menggunakannya) ketika saya berada di Surabaya tapi saya tidak jatuh ke bawah 0.} \\[-0.2ex] 
    & \color{teal}\texttt{Terjemahan: } \\[-0.1ex] 
    
    \textbf{Output:} & \texttt{beta memang pertama nae di Surabaya tapi son jato 0.} \\ 
    \bottomrule
    \end{tabular}
    }
    \caption{Example of Context-Based Instruction Data}
    \label{tab:contoh_prompt_context}

\end{table*}

\begin{table*}[ht!]
    \setlength{\tabcolsep}{6pt}
    \centering
    \footnotesize
    \resizebox{\linewidth}{!}{
    \begin{tabular}{r p{0.84\linewidth}}
    \toprule
    \textbf{Input:} & \color{olive} \texttt{Sinonim dalam bahasa Indonesia dan Melayu Kupang diberikan sebagai berikut:} \\[-0.8ex]
    & \color{purple}\texttt{- peras:pres} \\[-0.8ex]
    & \color{purple}\texttt{- lebih:lebe} \\[-0.8ex]
    & \color{purple}\texttt{- mentah:bahati} \\[-0.2ex]
    & \color{teal}\texttt{Jika teks dalam bahasa Indonesia adalah:} \color{violet}\texttt{Apa kau bisa memakan tiram mentah?,} \\[-0.2ex]
    & \color{teal}\texttt{apa artinya dalam bahasa Melayu Kupang?} \\[-0.1ex]
    
    \textbf{Output:} & \texttt{Lu bisa makan kerang manta ko?} \\
    \bottomrule
    \end{tabular}
    }
    \caption{Example of Semantics-Based Instruction Data}
    \label{tab:contoh_prompt_synonym}

\end{table*}

\begin{table*}[ht!]
    \setlength{\tabcolsep}{6pt}
    \centering
    \footnotesize
    \resizebox{\linewidth}{!}{
    \begin{tabular}{r p{0.84\linewidth}}
    \toprule
    \textbf{Input:} & \color{olive} \texttt{Perhatikan kalimat berikut dalam bahasa Melayu Kupang:} \\[-0.8ex]
    & \color{purple}\texttt{Oma bo'I pung gigi dong su rongko buang samua.} \\[-0.8ex]
    & \color{purple}\texttt{Kalo dia mau makan siri-pinang dia musti batumbuk dia pung siri-pinang deng hahabok} \\[-0.2ex]
    & \color{teal}\texttt{Teks ini memiliki makna penting dalam bahasa Indonesia:} \color{violet}\texttt{Karena seperti biasa tempat seperti itu adalah suci, kan?} \\[-0.2ex]
    & \color{teal}\texttt{Tolong terjemahkan ke bahasa Melayu Kupang.} \\[-0.1ex]

    \textbf{Output:} & \texttt{biasa kalo tempat-tempat begitu tu yang kramat a?} \\
    \bottomrule
    \end{tabular}
    }
    \caption{Example of Phonetic Instruction Data}
    \label{tab:contoh_prompt_makna}
   
\end{table*}

\begin{figure*}
    \centering
    \begin{minipage}{\linewidth}
        \resizebox{\linewidth}{!}{
            \includegraphics{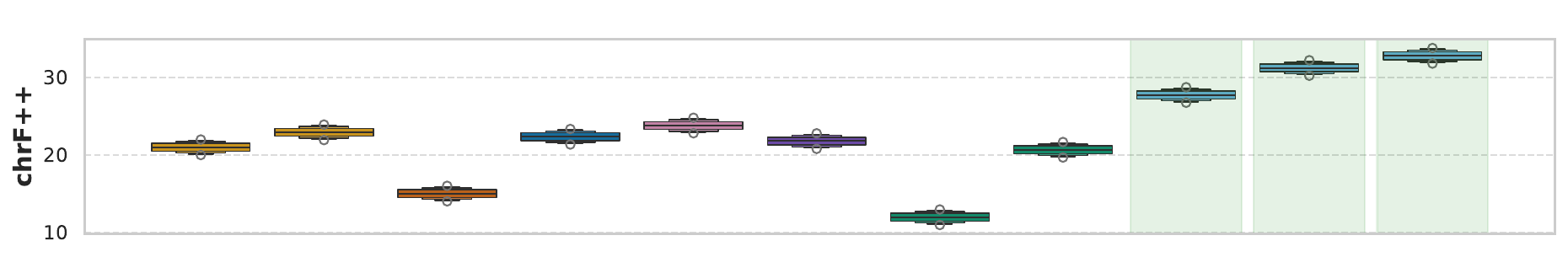}
        }
    \end{minipage}
    \begin{minipage}{\linewidth}
        \resizebox{\linewidth}{!}{
            \includegraphics{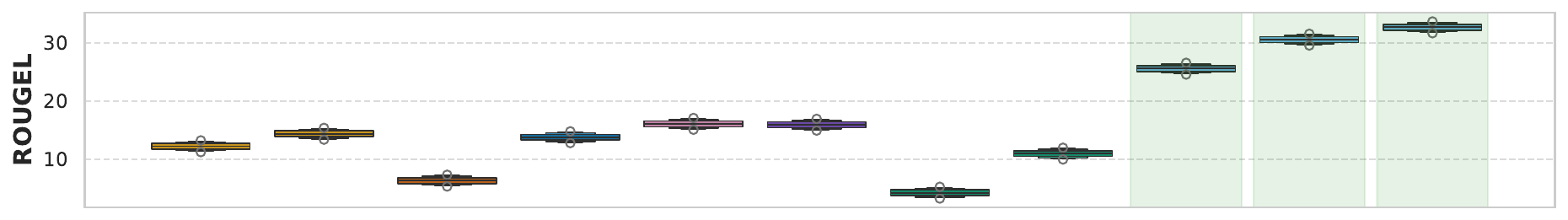}
        }
    \end{minipage}
    \begin{minipage}{\linewidth}
        \resizebox{\linewidth}{!}{
            \includegraphics[trim=0 0.75em 0 0,width=\textwidth,clip]{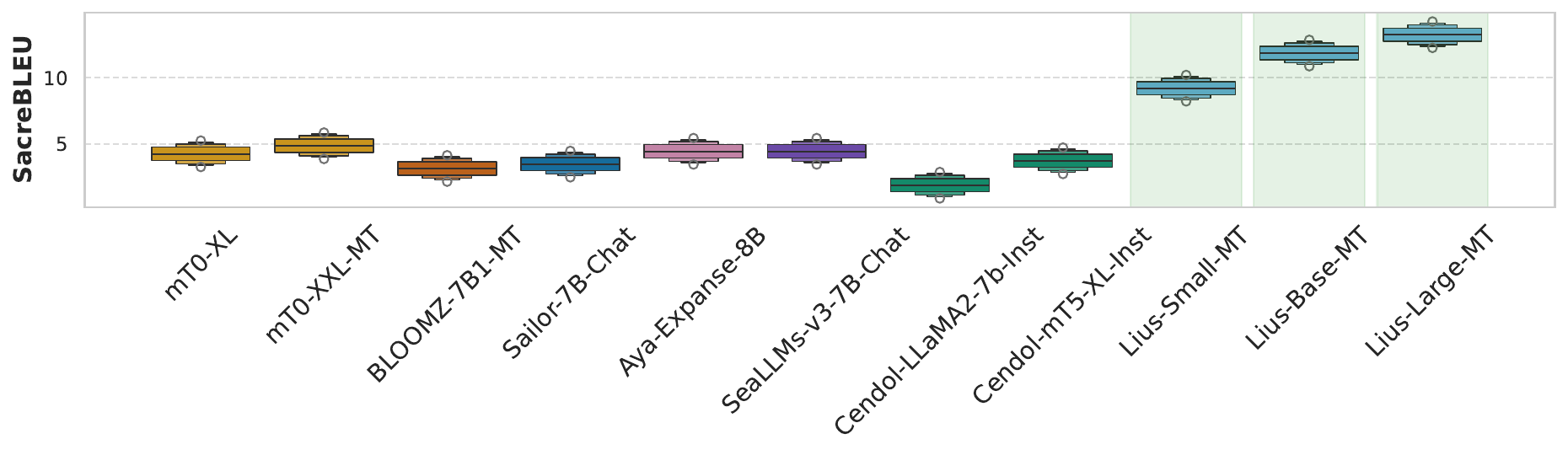}
        }        
    \end{minipage}
    \caption{Model Performance on Zero-shot Prompting}
    \label{fig:zero-shot}
\end{figure*}

\begin{figure*}
    \resizebox{\linewidth}{!}{
        \begin{minipage}{0.3\linewidth}
            \resizebox{\linewidth}{!}{
                \includegraphics[trim=0 0 0 0,clip]{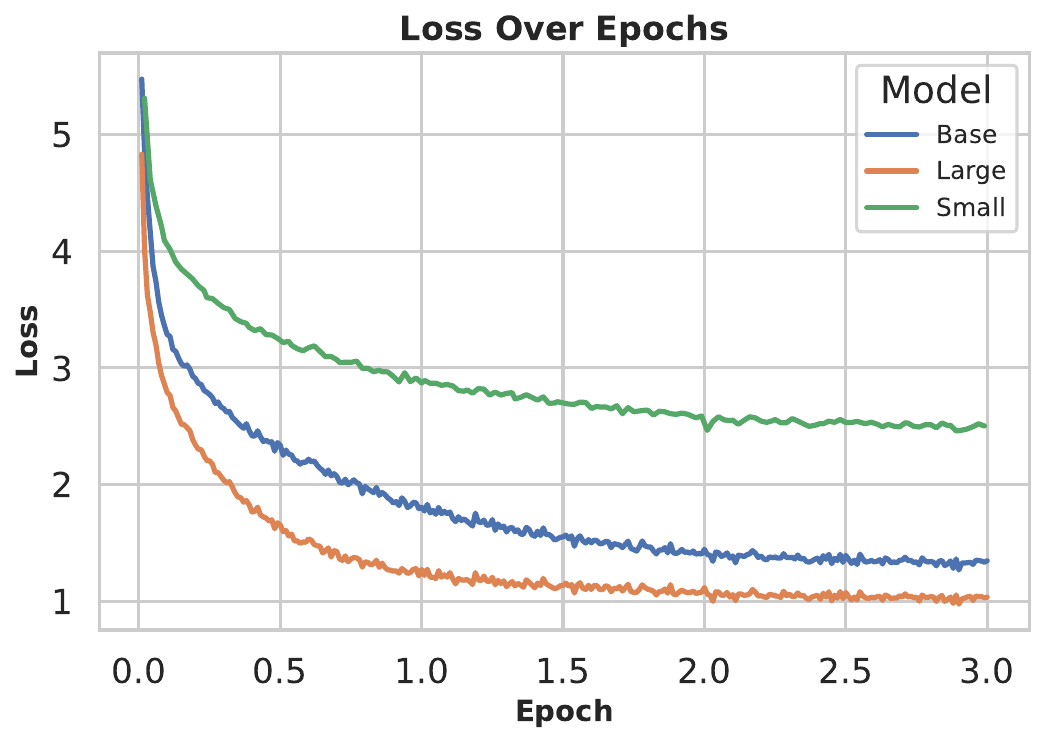}
            }
        \end{minipage}
        \hspace{2pt}
        \begin{minipage}{0.3\linewidth}
            \resizebox{\linewidth}{!}{
                \includegraphics[trim=0 0 0 0,clip]{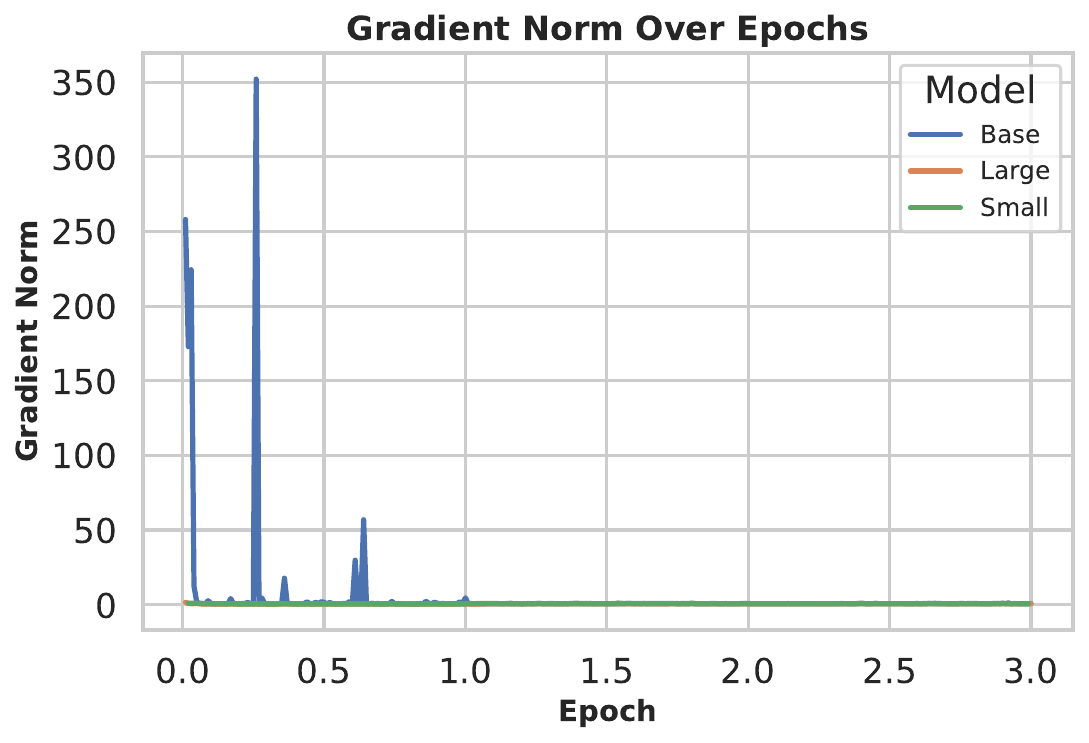}
            }
        \end{minipage}
        \hspace{2pt}
        \begin{minipage}{0.3\linewidth}
            \resizebox{\linewidth}{!}{
                \includegraphics[trim=0 0 0 0,clip]{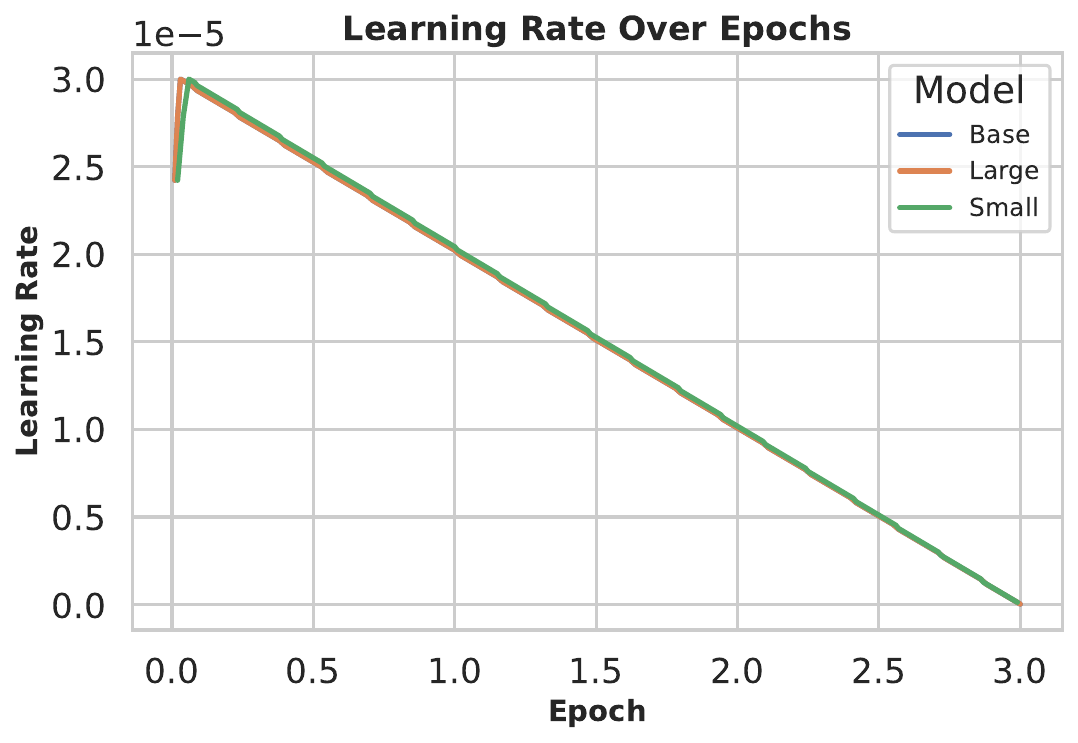}
            }
        \end{minipage}
    }
    \caption{Training Evaluation across Lius Model Variants}
    \label{fig:model_comparison}
\end{figure*}

\end{document}